\documentclass{aifrontiers}
\usepackage{aifrontiers}
\usepackage[T1]{fontenc}
\usepackage{booktabs}
\usepackage{amsfonts}
\usepackage{nicefrac}
\usepackage{array}
\usepackage{tabularx}
\usepackage{colortbl}
\usepackage{caption}
\usepackage{subcaption}
\usepackage{multirow}
\usepackage{multicol}
\usepackage{float}
\usepackage{wrapfig}
\usepackage{needspace}
\usepackage{placeins}
\usepackage{enumitem}
\usepackage{xspace}
\usepackage{soul}
\usepackage{listings}
\usepackage{algorithm}
\usepackage{algorithmic}

\AtBeginEnvironment{thebibliography}{\raggedright}

\newcommand{\tabref}[1]{Table~\ref{#1}}
\newcommand{\figref}[1]{Figure~\ref{#1}}
\newcommand{\secref}[1]{Section~\ref{#1}}

\renewcommand{\cite}{\citep}

\newlength\savewidth

\newcolumntype{x}[1]{>{\centering\arraybackslash}p{#1pt}}
\newcolumntype{y}[1]{>{\raggedright\arraybackslash}p{#1pt}}

\definecolor{baselinecolor}{gray}{0.93}

\newcommand{\addcite}[1]{}
\newcommand{\todo}[1]{}
\newcommand{\figplace}[3]{}
\newcommand{\figwide}[3]{}

\newcommand{\benchmark}{PersonaScienceBench\xspace}

\title{PersonaScienceBench: Evaluating Scientific Agents \\ with Diverse Simulated Researchers}
\shorttitle{PersonaScienceBench}
\author{
  {
    \mbox{\textbf{Siyang Wu$^{1}$}},
    \mbox{\textbf{Weizhi Zhang$^{2}$}},
    \mbox{\textbf{Chen Yang$^{3}$}},
    \mbox{\textbf{Esther Xu Fei}},
    \mbox{\textbf{Jorin Overwiening$^{3}$}},
    \mbox{\textbf{Xu Pan$^{3}$}},
    \mbox{\textbf{Jing Tang$^{4}$}},
    \mbox{\textbf{Yixuan He$^{5,\ddagger}$}},
    \mbox{Kelvin C. Mo$^{3}$},
    \mbox{Peiwen Li$^{6}$},
    \mbox{Defu Cao$^{7}$},
    \mbox{Xu Hu$^{8}$},
    \mbox{Haixin Wang$^{9}$},
    \mbox{Taoran Li$^{10}$},
    \mbox{Haoming Li$^{8}$},
    \mbox{Zheyuan Deng$^{11}$},
    \mbox{Jingwen Zhang$^{3}$},
    \mbox{Pengsong Zhang$^{12}$},
    \mbox{Haichao Zhang$^{13}$},
    \mbox{Guangze Luo$^{3}$},
    \mbox{Yi Daniel Lu$^{3}$},
    \mbox{Zihuai He$^{14,\ddagger}$},
    \mbox{Chenxi Qiu$^{3,\ddagger}$},
    \mbox{Dianzhuo Wang$^{3}$},
    \mbox{Zhejian Peng$^{15}$},
    \mbox{Shushu Liang$^{3}$},
    \mbox{Wenyi Lu$^{11}$},
    \mbox{Yifan Wang$^{16}$},
    \mbox{Jintao Huang$^{17}$},
    \mbox{Yijun Wang$^{3}$},
    \mbox{Wanting Jiang$^{18}$},
    \mbox{Yueying Li$^{19}$},
    \mbox{Robert George$^{20}$},
    \mbox{Haodi Hu$^{7}$},
    \mbox{Cong Wang$^{3}$},
    \mbox{Xiaomin Li$^{21,*}$},
    \mbox{Yuexing Hao$^{21,*}$}}\\[0.6em]
  {\small
    \mbox{$^{1}$University of Chicago} \quad
    \mbox{$^{2}$University of Illinois Chicago} \quad
    \mbox{$^{3}$Harvard University} \quad
    \mbox{$^{4}$Boston University} \quad
    \mbox{$^{5}$Arizona State University} \quad
    \mbox{$^{6}$Yale University} \quad
    \mbox{$^{7}$University of Southern California} \quad
    \mbox{$^{8}$The University of Texas at Dallas} \quad
    \mbox{$^{9}$University of California, Los Angeles} \quad
    \mbox{$^{10}$Texas A\&M University} \quad
    \mbox{$^{11}$Brown University} \quad
    \mbox{$^{12}$Department of Mechanical and Industrial Engineering, University of Toronto} \quad
    \mbox{$^{13}$Northeastern University} \quad
    \mbox{$^{14}$Departments of Neurology and Medicine, Stanford University} \quad
    \mbox{$^{15}$University of Illinois Urbana-Champaign} \quad
    \mbox{$^{16}$Purdue University} \quad
    \mbox{$^{17}$The Ohio State University} \quad
    \mbox{$^{18}$Amherst College} \quad
    \mbox{$^{19}$Massachusetts Institute of Technology} \quad
    \mbox{$^{20}$California Institute of Technology} \quad
    \mbox{$^{21}$MatrAIx}\\[0.3em]
    Bold: Core Team. \quad $^{\ddagger}$Faculty. \quad
    $^{*}$Corresponding Authors: \email{xiaominli@g.harvard.edu}, \email{yuexing@mit.edu}}
}
\date{\today}

\renewcommand{\weblink}{}
\renewcommand{\foundrylink}{}
\renewcommand{\hflink}{}
\renewcommand{\ghlink}{https://github.com/MatrAIx-ai/PersonaScienceBench/tree/public-release}
\renewcommand{\githubicon}{\raisebox{-1.5pt}{\includegraphics[height=1.2em]{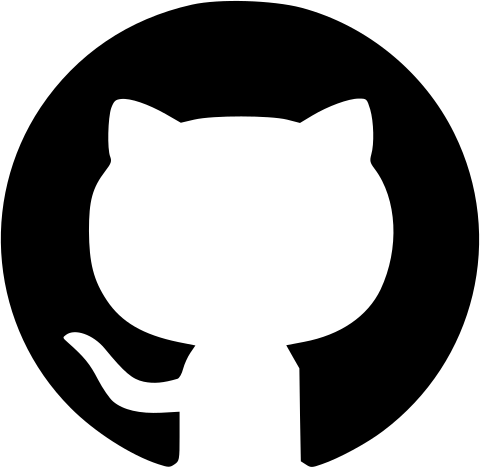}}}
\renewcommand{\AIFRenderLinksContent}{%
  \begin{center}
    \small\githubicon\ \textbf{Code:}\quad
    \href{\ghlink}{\nolinkurl{github.com/MatrAIx-ai/PersonaScienceBench/tree/public-release}}
  \end{center}
}

\begin{document}

\begin{abstract}
Scientific agents collaborate with researchers whose expertise and preferences
differ, but benchmarks usually hold the collaborator fixed. We introduce
\benchmark: 200 tasks from five benchmarks across four scientific domains.
Simulated researchers provide context and feedback according to persona
profiles, while task objectives and grading remain fixed. Agents encounter
researcher preferences only through conversation. Across twenty-three model
configurations, task performance, fact recovery, and self-repetition reveal
different strengths. Detailed GPT-6 Astra analyses show distinct agent
responses across researcher profiles; interaction traces illustrate
targeted clarification and workflow revision. These
behaviors can differ even when final scores match. \benchmark therefore
evaluates how agents collaborate with diverse researchers alongside whether
they complete the task.

\end{abstract}
\maketitle

\section{Introduction}
\label{sec:introduction}

\paragraph{Scientific work is collaborative.}
Language-model agents now perform substantial parts of the scientific
workflow. They search literature, propose hypotheses, write and execute
analysis code, and assemble results into reports
\cite{lu2024aiscientist,yamada2025aiscientistv2,gottweis2025coscientist,mitchener2025kosmos}.
The benchmarks used to measure this progress have followed the same
trajectory, moving from recall and reasoning questions toward executable
analysis tasks drawn from real studies
\cite{majumder2024discoverybench,mitchener2025bixbench,kon2025expbench,ye2025replicationbench}.
These capabilities do not remove the need for human collaboration, and
recent work questions their suitability for unsupervised discovery
\cite{bisht2026notbuilt}. Researchers often hold context that determines
whether an analysis is valid: why samples were excluded, whether a
covariate is confounded with batch, or which normalization a laboratory
uses. Such information may be absent from the data directory. An agent's
scientific output can therefore depend on how it consults a researcher
who may explain readily, answer only the literal question, or misremember
a detail.

\begin{figure}[!htbp]
\centering
\includegraphics[width=\linewidth]{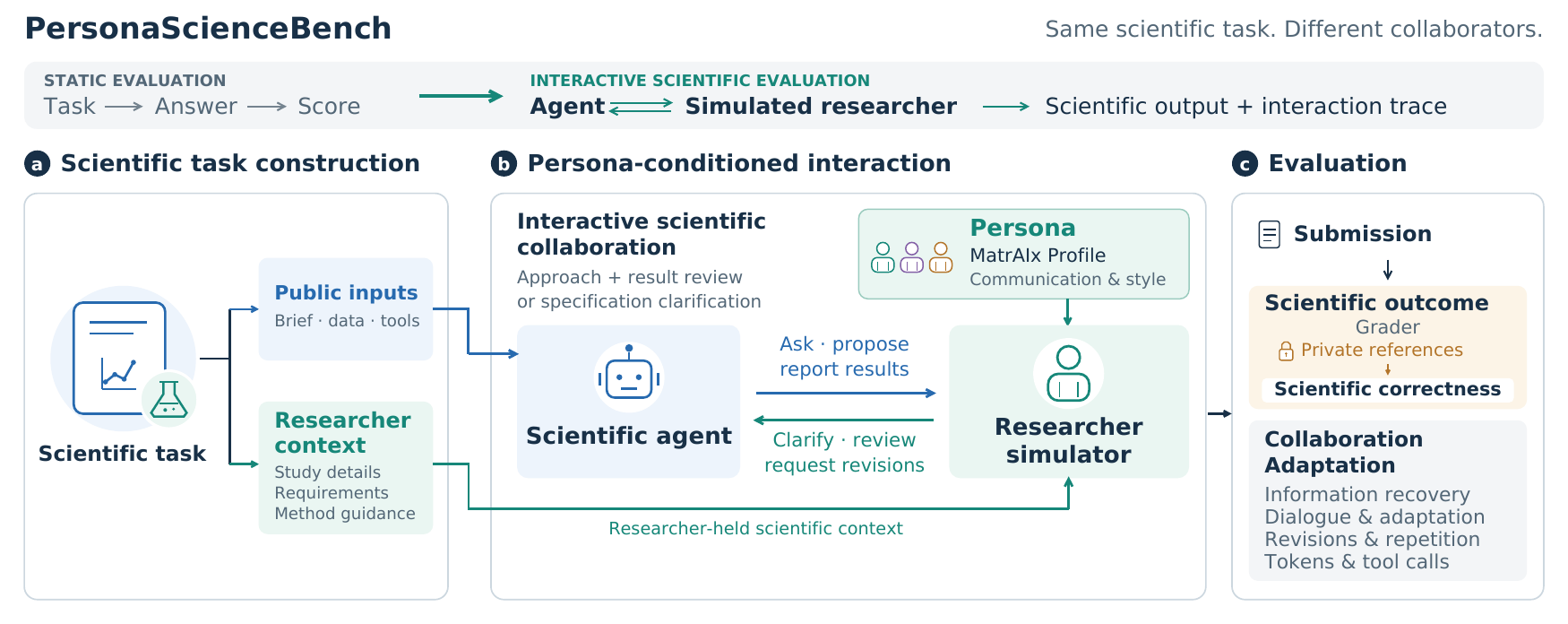}
\caption{\textbf{PersonaScienceBench overview.}
\textbf{a,} Separate public inputs from researcher-held context.
\textbf{b,} Interact with a persona-conditioned researcher.
\textbf{c,} Evaluate scientific outcomes, collaboration, and adaptation.}
\label{fig:workflow}
\end{figure}

\begin{wrapfigure}{r}{0.48\textwidth}
\vspace{-\intextsep}
\centering
\includegraphics[width=\linewidth]{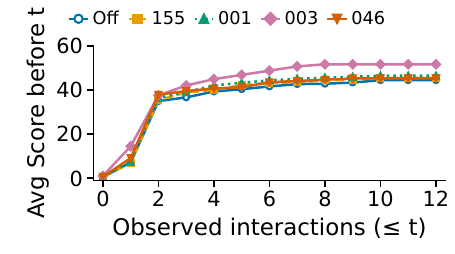}
\caption{\textbf{Performance by interaction length.}
Cumulative final scores from episodes ending by $t$, divided by the
200 task episodes in each setting. Curves describe completion
lengths, not within-run learning.}
\label{fig:interaction-performance}
\vspace{-6pt}
\end{wrapfigure}

\paragraph{The evaluation gap.}
Scientific agent benchmarks typically provide complete task specifications.
Interactive benchmarks instead test escalation, clarification, and
tool-agent interaction
\cite{trinh2026hilbench,somasekharan2026sciconvbench,yao2024taubench,chen2026appworldul,ta2026regretbench},
but usually hold the collaborator fixed. Non-collaborative simulators and
persona policies vary user behavior in service domains
\cite{shim2025noncollaborative,chopra2026ppol}; they do not test its
consequences for scientific correctness. Varying the researcher while holding the task and agent fixed makes the
collaborator's influence observable. Figure~\ref{fig:interaction-performance}
illustrates why interaction also matters: episodes contribute to final
performance at different completion lengths across researcher profiles.
Evaluation should therefore capture how agents clarify requirements,
respond to feedback, and revise their work alongside whether they succeed.

\paragraph{Our approach.}
We introduce \benchmark, which varies the simulated researcher on fixed
scientific tasks (\figref{fig:workflow}). We convert 200 tasks from
DiscoveryBench~\cite{majumder2024discoverybench},
BixBench~\cite{mitchener2025bixbench}, EXP-Bench~\cite{kon2025expbench},
SciConvBench~\cite{somasekharan2026sciconvbench}, and
ASI-Bench~\cite{zhou2026asibench} into collaborative tasks with clarification or review while preserving source-specific deliverables and
grading. The researcher receives a complete rendered MatrAIx profile
\cite{li2026matraix}: an auditable attribute--value listing, typically
containing roughly 200 populated attributes from a schema of 1,290
categorical dimensions. Only the researcher simulator receives this
identity block; the actor observes researcher preferences through conversation, and evaluator
references remain private. In the persona-off condition, the simulator receives no persona profile;
all other instructions remain unchanged. Section~\ref{sec:statistical-protocol} specifies
the trial protocol and uncertainty estimates for these comparisons.

\paragraph{What we measure.}
We assess scientific outcomes and interaction behavior separately
(\figref{fig:workflow}c). Source-specific graders measure task performance;
interaction metrics capture message length, tool use, clarification,
fact recovery, and self-repetition. Together, they distinguish
successful completion from intermediate progress, such as recovering
missing information in an episode that still ends unsuccessfully. They
also reveal changes in collaboration that a final score alone can miss.

\paragraph{Contributions.}
Our contributions are threefold:
\begin{itemize}[leftmargin=*,nosep]
    \item \textbf{A benchmark for scientific collaboration with diverse researchers.}
    We introduce \benchmark, adapting 200 tasks from five benchmarks across
    four domains into interactions involving clarification, review, and
    revision. Researcher personas vary while task objectives and grading
    criteria remain fixed. Only the simulator receives the profile; the actor
    encounters researcher preferences through interaction.
    \item \textbf{Evidence of persona diversity and agent adaptation.}
    We show distinct simulator and actor responses and interaction
    trajectories across researcher personas. Behavioral analyses and case
    studies reveal differences in communication volume, targeted follow-up
    questions, and workflow revision.
    \item \textbf{Evaluation beyond final scores.}
    We evaluate twenty-three model configurations using scientific performance,
    fact recovery, and self-repetition. Detailed persona analyses show how
    different collaboration processes and information-gathering progress
    can be obscured by final scores, making these behaviors observable
    alongside task outcomes.
\end{itemize}

\section{Related Work}
\label{sec:related-work}

\noindent\textbf{Language Models with Personas.}
Conditioning language models on persona descriptions can produce varied
responses and simulated user behavior
\cite{argyle2022outofone,ge2024personahub}. Recent work develops persona
generators for user coverage and applies persona-based simulation to
interactive-system evaluation
\cite{paglieri2026personagenerators,liu2026personaeval}.
Non-collaborative user simulators and Persona Policies further show that
varying user behavior can change tool-agent performance in service domains
\cite{shim2025noncollaborative,chopra2026ppol}. MatrAIx provides structured
profiles spanning background, capability, and behavior
\cite{li2026matraix}. We use these profiles to study simulated scientific
collaborators, comparing persona-conditioned interactions with a persona-off
control while holding scientific tasks and interaction protocols fixed.

\noindent\textbf{Interactive Agent Evaluation.}
Interactive benchmarks increasingly test whether agents ask for missing
information and adapt to user feedback. $\tau$-bench evaluates tool-agent-user
interaction under domain rules \cite{yao2024taubench}; HiL-Bench targets
escalation and help-seeking \cite{trinh2026hilbench}; AppWorld-UL and
regret-based clarification benchmarks study multi-turn interaction and
underspecification \cite{chen2026appworldul,ta2026regretbench}. These settings
establish that interaction policy is itself a capability, but they generally
hold the collaborator fixed or study non-scientific service domains.
\benchmark instead varies the researcher while retaining a graded scientific
objective, allowing collaborator-conditioned behavior and scientific outcome
to be analyzed together.

\noindent\textbf{Agents for Scientific Discovery.}
Language-model agents increasingly support hypothesis generation,
experimental implementation, and data analysis, as illustrated by the AI
Scientist, AI co-scientist, and Kosmos
\cite{lu2024aiscientist,yamada2025aiscientistv2,gottweis2025coscientist,mitchener2025kosmos}.
Benchmarks evaluate these capabilities through data-driven discovery and
computational biology
\cite{majumder2024discoverybench,mitchener2025bixbench}, experimental
design and execution \cite{kon2025expbench}, and scientific implementation
under varying methodological guidance \cite{zhou2026asibench}.
SciConvBench extends evaluation to multi-turn clarification of
underspecified computational-science requests
\cite{somasekharan2026sciconvbench}. These benchmarks assess scientific
task performance and information recovery, but generally leave collaborator
behavior fixed. \benchmark builds on their task settings by varying the
simulated researcher on matched scientific tasks and measuring interaction
patterns alongside source-specific outcomes. This design makes differences in the collaborative process visible
alongside differences in scientific outcomes.

\section{PersonaScienceBench}
\label{sec:benchmark}

\subsection{Persona Construction}
\label{sec:persona-construction}
\paragraph{Source personas.}
We draw simulated researcher profiles from the public Persona8B coreset
released by MatrAIx~\cite{li2026matraix}. Its underlying population uses a schema of 1,290 categorical dimensions spanning background, psychology, capability, interaction, and lifestyle. MatrAIx constructs records through two pathways. Synthetic records are sampled in topological order from a directed acyclic dependency graph: source-informed categorical priors and conditional adjustments preserve selected correlations, while compatibility rules exclude specified contradictions. Human-grounded records map
biographies, review histories, coded surveys, and consented self-reports into the same schema; fields without source support remain unassigned. Both paths undergo source-appropriate quality checks and deduplication. The released coreset contains 599,847 human-grounded and 400,000 synthetic records. This construction provides varied, inspectable identity prompts; it does not establish that a simulated collaborator represents a particular real researcher or a population subgroup.

\paragraph{From records to simulated researchers.}
We select four coreset profiles using a coverage-based strategy over
communication, decision-making, interaction, tool-collaboration, and workflow
attributes. Together with persona off, they define the five
settings used throughout the evaluation. For each selected record, the pinned MatrAIx renderer emits every retained populated
attribute as a category-grouped attribute--value identity block; null,
placeholder, schema-default, and designated source-dump fields are omitted.
We inject this complete rendered block into the researcher simulator's system
prompt, without task-specific pruning or summarization. The actor and
evaluator never receive it. The persona-off condition uses the same simulator
and interaction protocol with the identity block removed. Scientific task
facts and grading criteria are held fixed across conditions (\secref{sec:dataset-construction}).

\tabref{tab:focal-personas} summarizes the backgrounds and interaction
preferences of the four profiles used in the detailed persona analyses.
The same four profiles are used for every model and task. Their comparisons
characterize sensitivity to these collaborator settings, rather than an
average effect over a population of researchers. The case studies in
Section~\ref{sec:discussion} and Appendix~\ref{app:discussion-interactions}
additionally use persona 127, an English-language coreset profile with a
background in education and curriculum design
(Appendix~\ref{app:persona-profile-127}). It was run with GPT-6 Astra on the
same tasks, under the same protocol and against the same persona-off
conversations, so its examples need no translation. It is not one of the five
evaluation settings and does not enter any reported result.

\begin{table}[htbp]
  \centering
  \begingroup
  \footnotesize
  \setstretch{1.0}
  \setlength{\tabcolsep}{5pt}
  \renewcommand{\arraystretch}{1.2}
  \begin{tabularx}{\linewidth}{@{}l>{\raggedright\arraybackslash}Xl>{\raggedright\arraybackslash}X@{}}
    \toprule
    Persona & Background and specialty & Primary language & Interaction preferences \\
    \midrule
    001 & Business and management; curriculum design & Hindi
        & Brief, direct communication. \\
    \addlinespace[3pt]
    003 & Engineering; electrical work & Russian
        & Detailed, concrete explanations; speed and novel approaches. \\
    \addlinespace[3pt]
    046 & Finance and economics; machine learning & Spanish
        & Formal, indirect communication. \\
    \addlinespace[3pt]
    155 & Agriculture; agronomy & Mandarin
        & Detailed guidance; accuracy, familiar methods, and teamwork. \\
    \bottomrule
  \end{tabularx}
  \endgroup
  \caption{Backgrounds and interaction preferences of the four researcher personas.}
  \label{tab:focal-personas}
\end{table}

\subsection{From Static Benchmarks to Interactive Simulations}
\label{sec:dataset-construction}

We adapt scientific tasks so that an actor must work with a simulated
researcher, exposing decisions that final-answer evaluation alone does not
capture. Conversion separates public inputs, researcher context, and private
evaluation, and defines when the agents interact to share information
(Algorithm~\ref{alg:benchmark-episode}). The aim is to make consultation and
review observable while preserving a well-defined scientific task. Required
checkpoints are distinguished from voluntary clarification and revision;
longer conversations are not rewarded.

The benchmark contains 200 tasks, with 40 each from DiscoveryBench
\cite{majumder2024discoverybench}, BixBench \cite{mitchener2025bixbench},
EXP-Bench \cite{kon2025expbench}, SciConvBench
\cite{somasekharan2026sciconvbench}, and ASI-Bench \cite{zhou2026asibench}.
\tabref{tab:dataset-construction} gives their distribution across four
domains. The selection covers numerical scientific workflows and diagnostic
plots; tasks involving image processing are outside the scope of this study.
\par\medskip\noindent
\begin{minipage}{\linewidth}
  \begin{minipage}[t]{0.45\linewidth}
    \vspace{0pt}
    \captionsetup{justification=raggedright,singlelinecheck=false,hypcap=false}
    \begingroup
    \footnotesize
    \setstretch{1.0}
    \setlength{\tabcolsep}{3pt}
    \renewcommand{\arraystretch}{1.15}
    \begin{tabular*}{\linewidth}{@{\extracolsep{\fill}}lrrrrr@{}}
      \toprule
      Benchmark & Life & \shortstack{Phys./\\Eng.}
        & \shortstack{Math./\\CS} & \shortstack{Soc./\\Hum.} & Total \\
      \midrule
      DiscoveryBench & 5 & 0 & 3 & 32 & 40 \\
      BixBench & 40 & 0 & 0 & 0 & 40 \\
      EXP-Bench & 0 & 6 & 34 & 0 & 40 \\
      SciConvBench & 0 & 34 & 6 & 0 & 40 \\
      ASI-Bench & 5 & 23 & 12 & 0 & 40 \\
      \midrule
      \textbf{Total} & \textbf{50} & \textbf{63} & \textbf{55} & \textbf{32} & \textbf{200} \\
      \bottomrule
    \end{tabular*}
    \par\endgroup
    \captionof{table}{Task counts by benchmark and domain.
    Domains: Life, life sciences; Phys./Eng., physical sciences
    and engineering; Math./CS, mathematics and computer science;
    Soc./Hum., social sciences and humanities.}
    \label{tab:dataset-construction}
  \end{minipage}\hfill
  \begin{minipage}[t]{0.52\linewidth}
    \vspace{0pt}
    \captionsetup{justification=raggedright,singlelinecheck=false,hypcap=false}
    \footnotesize
    \setstretch{1.0}
    \hrule\smallskip
    \begin{algorithmic}[1]
      \REQUIRE Source task $t$; persona $p$ (or persona off)
      \STATE $(b,k,e,\pi) \gets \mathrm{Convert}(t)$
      \STATE Initialize actor $A$ with public inputs $b$.
      \STATE Initialize simulator $S$ with context $k$ and $p$.
      \STATE $h \gets []$ \COMMENT{shared conversation}
      \WHILE{protocol $\pi$ permits another turn}
        \IF{$\pi$ assigns the next turn to $A$}
          \STATE $m \gets A.\mathrm{ActAndReport}(h)$
        \ELSE
          \STATE $m \gets S.\mathrm{Respond}(h)$
        \ENDIF
        \STATE Append $m$ to $h$.
      \ENDWHILE
      \STATE $y \gets \mathrm{SealSubmission}(A,h)$
      \RETURN $e.\mathrm{Evaluate}(y,h)$
    \end{algorithmic}
    \smallskip\hrule
    \captionof{algorithm}{Task conversion and interactive simulation.
    Conversion separates public inputs $b$, researcher context $k$, and
    private evaluator $e$, and selects protocol $\pi$. Actor actions use
    only the tools permitted by the task; evaluator references stay private.}
    \label{alg:benchmark-episode}
  \end{minipage}
\end{minipage}
\par\medskip

\noindent DiscoveryBench, BixBench, EXP-Bench, and ASI-Bench use approach and
result reviews: the researcher approves a proposed approach, then requests
an identifiable revision or concludes after reviewing the reported result.
SciConvBench retains its native specification-clarification dialogue.
Evaluator references remain private, and only the researcher receives the
persona profile. Source-specific deliverables and grading are retained
within declared adaptations, including EXP-Bench's design-only track and
ASI-Bench's objective-only actor with researcher-held methodological
guidance. Appendix~\ref{app:conversion-principles} details the shared
conversion rules, information boundaries, interaction protocol, and
acceptance rubric; Appendices~\ref{app:discoverybench}--\ref{app:asibench} describe the
benchmark-specific implementations.

\section{Experimental Setup}
\label{sec:experimental-setup}

\subsection{Models, Persona Settings, and Evaluation Protocol}
\label{sec:run-coverage}
\label{sec:statistical-protocol}
We evaluate twenty-three model configurations (\tabref{tab:model-scorecard}).
The Anthropic models are Claude Opus 5.5~\cite{anthropic2026opus55},
Claude Fable 5.1~\cite{anthropic2026fable51},
Claude Sonnet 5.5~\cite{anthropic2026sonnet55},
Claude Sonnet 5~\cite{anthropic2026sonnet5},
Claude Haiku 5.5~\cite{anthropic2026haiku55}, and
Claude Haiku 4.5~\cite{anthropic2025haiku45}. From Google DeepMind, we
evaluate Gemini 3.8 Flash~\cite{deepmind2026gemini38flash},
Gemini 3.7 Flash~\cite{deepmind2026gemini37flash}, and
Gemini 3 Flash~\cite{deepmind2025gemini3flash}. The OpenAI models are
GPT-6 Astra, GPT-6 Sol, and GPT-6 Luna~\cite{openai2026astra};
GPT-5.6 Terra~\cite{openai2026gpt56}; GPT-5.5~\cite{openai2026gpt55}; and
GPT-5 mini and GPT-5 nano~\cite{openai2025gpt5}. We also evaluate
Grok 4.6 from xAI~\cite{xai2026grok46}, Muse Glimmer 30B from
Meta~\cite{meta2026museglimmer}, GLM-5.3-Flash from Z.ai~\cite{zai2026glm53flash},
Qwen3.8-27B~\cite{qwen2026qwen38} and Qwen3.8-Flash~\cite{alibaba2026qwen38flash}
from Alibaba's Qwen team, DeepSeek-V4-Flash from DeepSeek~\cite{deepseek2026v4}, and
MiniMax-M3 from MiniMax~\cite{minimax2026m3}.
Each configuration uses the same model
for the actor and researcher simulator. All model-based evaluations use
GPT-6 Astra with low reasoning effort. GPT-6 Astra also uses low reasoning
effort when acting as the actor and simulator. The detailed persona analyses in
Sections~\ref{sec:persona-performance}--\ref{sec:interaction-outcomes}
use the set of runs with GPT-6 Astra.

We evaluate five settings: the four profiles in \tabref{tab:focal-personas}
and persona off. Only the researcher simulator receives the profile;
persona off omits it while retaining the same interaction protocol.
Every model is evaluated on the same 200 questions, comprising 40 from each
benchmark, with one trial per question and setting. Thus each model has
1,000 task--setting episodes. For GPT-6 Astra, persona off and personas 155
and 001 were each run up to three times, and Astra's performance estimates
average the available trials for each task (Appendix~\ref{app:analysis-coverage}). Tasks, scientific requirements, and grading criteria remain fixed across settings.

For Astra performance and self-repetition, 95\% confidence intervals
use 20,000 study-cluster
bootstrap resamples within benchmarks~\cite{field2007bootstrap}, with shared
resampling draws across settings. These intervals describe variation across
sampled studies, conditional on the recorded trials; they do not
estimate variability across repeated rollouts.
Appendix~\ref{app:analysis-coverage} details aggregation and metric eligibility.

\subsection{Evaluation Metrics}
\label{sec:metrics}

\noindent\textbf{Scientific performance.}
We normalize each benchmark's score to a 0--100 scale. Performance for setting $p$ is
$S_p=|\mathcal B|^{-1}\sum_{b\in\mathcal B}|Q_b|^{-1}\sum_{q\in Q_b}s_{qp}$,
where $s_{qp}$ is the normalized score of question $q$ under setting $p$
(averaged over available trials for GPT-6 Astra), $Q_b$ is the set of
questions from benchmark $b$, and $\mathcal B$ contains the five benchmarks.
The design uses the same 40 questions per benchmark for every model and
setting; Appendix~\ref{app:analysis-coverage} describes the questions available
to each estimate. Mean Persona Performance is
$S_{\mathrm{persona}}=|\mathcal P|^{-1}\sum_{p\in\mathcal P}S_p$ for
$\mathcal P=\{155,001,003,046\}$; persona off is reported separately.

\noindent\textbf{Fact recovery.}
For SciConvBench disambiguation cases, fact recovery is
$F_e=100\,|H_e\cap R_e|/|H_e|$, where $H_e$ is the nonempty set of required
withheld facts and $R_e$ is the set recovered during episode $e$.

\noindent\textbf{Actor self-repetition.}
We measure lexical reuse across actor messages using word windows.
For window $w$ in eligible message $m$, the match indicator is
$I_{em}(w)=\mathbf{1}\{\max_{v\in W_{e,<m}}\sum_{k=1}^{6}\mathbf{1}\{w_k=v_k\}\geq3\}$,
where $W_{e,<m}$ contains windows from earlier eligible actor messages and
$\mathbf{1}$ denotes the indicator function. Episode-level repetition is
$\operatorname{Rep}_e=\frac{100}{M_e-1}\sum_{m=2}^{M_e}|W_{em}|^{-1}\sum_{w\in W_{em}}I_{em}(w)$,
where $W_{em}$ contains the windows in message $m$ and $M_e\geq2$ is the
number of eligible messages. Messages must contain at least nine word
tokens, matching the fixed eligibility rule used in the sensitivity grid
(Appendix~\ref{app:analysis-coverage}).

\section{Results}
\label{sec:results}

\subsection{Model-Level Scientific Collaboration}
\label{sec:model-results}
\begin{table}[!t]
\centering
\begingroup
\footnotesize
\setstretch{1.0}
\ifdefined\tabciwidth\else\newlength{\tabciwidth}\fi
\settowidth{\tabciwidth}{\scriptsize[00.00, 00.00]}
\newcommand{\tabci}[1]{\,\makebox[\tabciwidth][l]{\scriptsize[#1]}}
\newcommand{\tabcinone}{\,\makebox[\tabciwidth]{}}
\setlength{\tabcolsep}{3pt}
\renewcommand{\arraystretch}{1.15}

\begin{tabular*}{\linewidth}{@{\extracolsep{\fill}}lcccc@{}}
\toprule
Actor/simulator model
& \shortstack{Persona-off\\performance}
& \shortstack{Mean Persona\\Performance}
& \shortstack{Fact\\recovery (\%)}
& \shortstack{Self-\\repetition (\%)} \\
\midrule
\multicolumn{5}{l}{
\makebox[1.5em][c]{\raisebox{-0.15em}{%
\includegraphics[height=1.05em,width=1.5em,keepaspectratio]{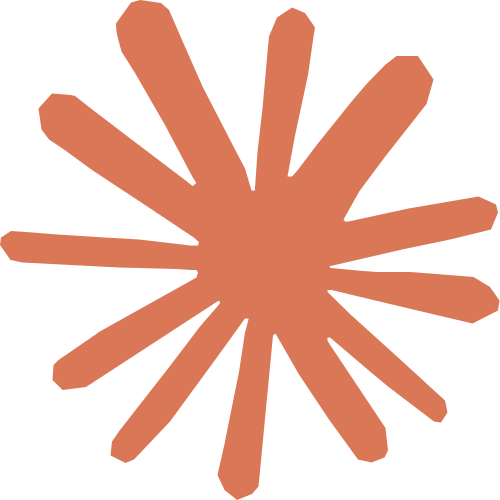}%
}}
\hspace{0.3em}
\textit{Anthropic Claude}
} \\

Claude Opus 5.5
& 49.48\tabci{45.02, 54.14}
& 50.55\tabci{48.15, 53.06}
& 98.95\tabci{97.83, 99.86}
& 16.23\tabci{14.79, 17.72}
\\

Claude Fable 5.1
& 51.00\tabci{44.42, 57.25}
& 47.09\tabci{40.74, 52.85}
& 93.92\tabci{83.71, 99.75}
& 15.33\tabci{14.41, 16.34}
\\

Claude Sonnet 5.5
& 45.31\tabci{40.61, 50.02}
& 44.72\tabci{42.22, 47.17}
& 95.17\tabci{91.36, 98.19}
& 18.86\tabci{16.84, 20.87}
\\

Claude Sonnet 5
& 34.48\tabci{29.82, 39.27}
& 34.30\tabci{31.94, 36.73}
& 89.83\tabci{84.37, 94.60}
& 24.91\tabci{23.38, 26.50}
\\

Claude Haiku 5.5
& 33.52\tabci{29.34, 37.76}
& 34.23\tabci{32.10, 36.44}
& 91.59\tabci{87.27, 95.36}
& 32.24\tabci{30.48, 34.01}
\\

Claude Haiku 4.5
& 21.97\tabci{16.25, 27.71}
& 19.72\tabci{15.19, 24.14}
& 67.95\tabci{61.77, 74.37}
& 21.43\tabci{17.23, 24.53}
\\

\addlinespace[2pt]

\multicolumn{5}{l}{
\makebox[1.5em][c]{\raisebox{-0.15em}{%
\includegraphics[height=1.05em,width=1.5em,keepaspectratio]{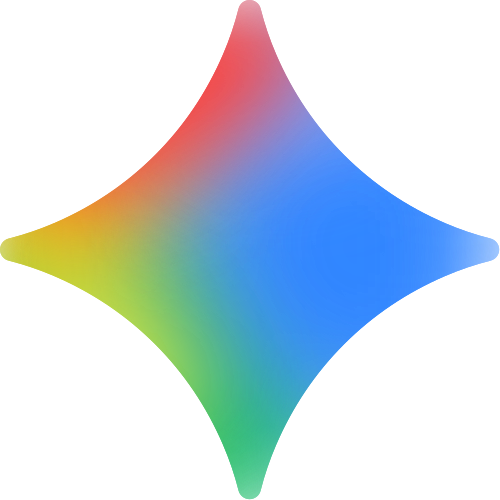}%
}}
\hspace{0.3em}
\textit{Google Gemini}
} \\

Gemini 3.8 Flash
& 36.37\tabci{30.09, 42.23}
& 36.42\tabci{30.72, 41.96}
& 81.48\tabci{67.05, 92.96}
& 21.77\tabci{20.12, 23.99}
\\

Gemini 3.7 Flash
& 48.21\tabci{42.73, 53.77}
& 48.17\tabci{45.56, 50.83}
& 75.24\tabci{67.27, 82.79}
& 13.08\tabci{11.05, 15.13}
\\

Gemini 3 Flash
& 33.07\tabci{28.55, 37.91}
& 31.98\tabci{29.46, 34.42}
& 90.01\tabci{85.65, 93.85}
& 13.43\tabci{11.72, 15.23}
\\

\addlinespace[2pt]

\multicolumn{5}{l}{
\makebox[1.5em][c]{\raisebox{-0.15em}{%
\includegraphics[height=1.05em,width=1.5em,keepaspectratio]{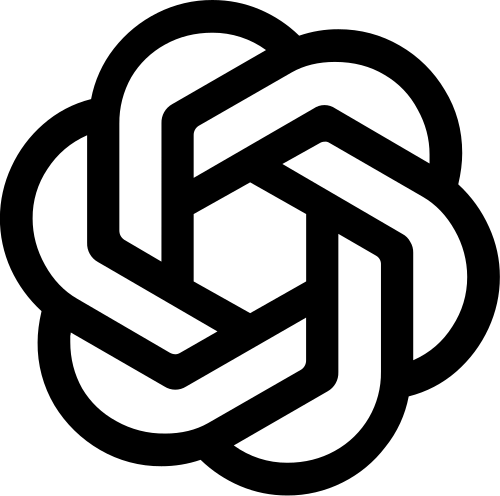}%
}}
\hspace{0.3em}
\textit{OpenAI GPT}
} \\

GPT-6 Astra
& 44.88\tabci{37.47, 51.95}
& 46.23\tabci{38.81, 53.30}
& 88.86\tabci{76.96, 97.68}
& 12.69\tabci{11.65, 13.73}
\\

GPT-6 Sol
& 38.52\tabci{33.28, 43.71}
& 38.29\tabci{35.78, 40.81}
& 78.42\tabci{71.70, 84.52}
& 17.16\tabci{14.84, 19.61}
\\

GPT-6 Luna
& 33.08\tabci{28.19, 38.03}
& 32.62\tabci{30.21, 35.06}
& 71.07\tabci{63.32, 78.65}
& 30.73\tabci{27.79, 33.75}
\\

GPT-5.6 Terra
& 34.56\tabci{30.05, 39.16}
& 33.33\tabci{30.95, 35.73}
& 83.56\tabci{76.70, 89.83}
& 12.41\tabci{10.54, 14.37}
\\

GPT-5.5
& 43.47\tabci{37.66, 49.48}
& 37.60\tabci{35.04, 40.22}
& 84.85\tabci{78.57, 90.39}
& 13.76\tabci{11.77, 15.82}
\\

GPT-5 mini
& 26.37\tabci{22.19, 30.94}
& 31.63\tabci{28.32, 35.10}
& 68.42\tabci{60.71, 75.99}
& 10.52\tabci{8.78, 12.33}
\\

GPT-5 nano
& 14.56\tabci{10.41, 18.69}
& 15.55\tabci{13.29, 17.83}
& 63.26\tabci{55.83, 70.76}
& 15.99\tabci{13.70, 18.44}
\\

\addlinespace[2pt]

\multicolumn{5}{l}{
\makebox[1.5em][c]{\raisebox{-0.15em}{%
\includegraphics[height=1.05em,width=1.5em,keepaspectratio]{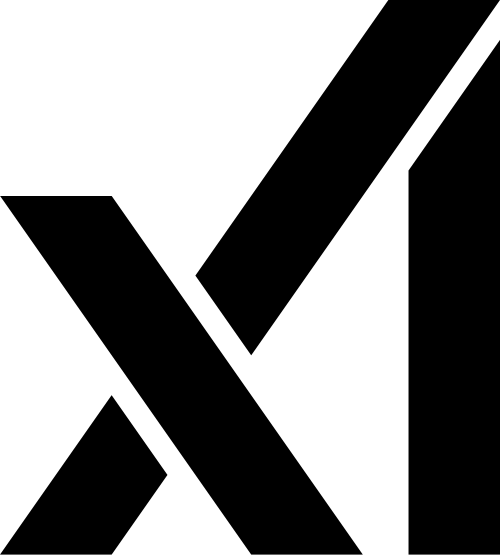}%
}}
\hspace{0.3em}
\textit{xAI Grok}
} \\

Grok 4.6
& 39.78\tabci{33.29, 45.99}
& 39.29\tabci{33.73, 44.45}
& 86.64\tabci{74.97, 95.31}
& 13.22\tabci{11.98, 14.58}
\\

\addlinespace[2pt]

\multicolumn{5}{l}{
\makebox[1.5em][c]{\raisebox{-0.15em}{%
\includegraphics[height=1.05em,width=1.5em,keepaspectratio]{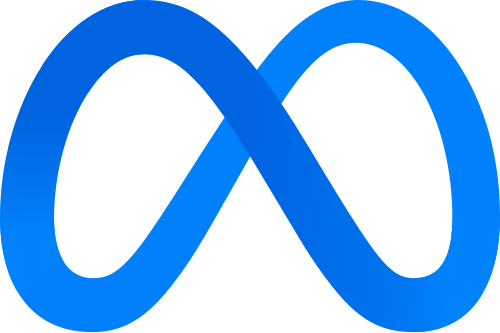}%
}}
\hspace{0.3em}
\textit{Meta Muse}
} \\

Muse Glimmer 30B
& 36.95\tabci{31.35, 42.39}
& 34.26\tabci{29.17, 39.23}
& 86.15\tabci{75.15, 95.00}
& 37.62\tabci{35.56, 39.94}
\\

\addlinespace[2pt]

\multicolumn{5}{l}{
\makebox[1.5em][c]{\raisebox{-0.15em}{%
\includegraphics[height=1.05em,width=1.5em,keepaspectratio]{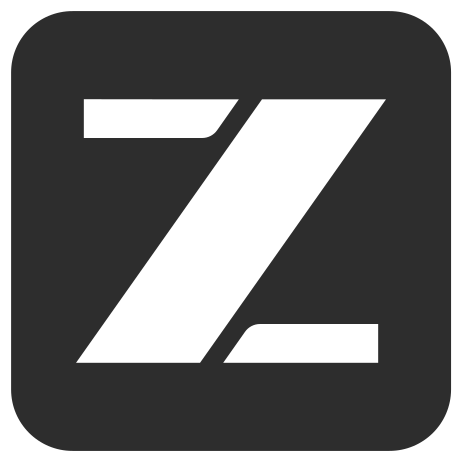}%
}}
\hspace{0.3em}
\textit{Z.ai GLM}
} \\

GLM-5.3-Flash
& 33.34\tabci{27.20, 39.17}
& 32.13\tabci{26.28, 37.50}
& 84.61\tabci{72.07, 94.73}
& 17.19\tabci{15.70, 18.90}
\\

\addlinespace[2pt]

\multicolumn{5}{l}{
\makebox[1.5em][c]{\raisebox{-0.15em}{%
\includegraphics[height=1.05em,width=1.5em,keepaspectratio]{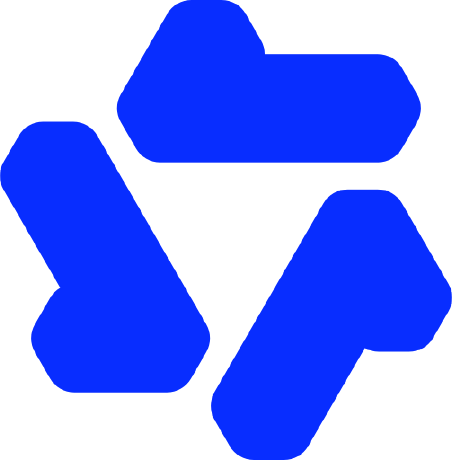}%
}}
\hspace{0.3em}
\textit{Alibaba Qwen}
} \\

Qwen3.8-27B
& 32.73\tabci{26.62, 38.66}
& 30.66\tabci{25.08, 35.88}
& 84.14\tabci{69.26, 96.13}
& 25.18\tabci{22.68, 28.36}
\\

Qwen3.8-Flash
& 32.32\tabci{26.10, 38.77}
& 30.51\tabci{25.92, 35.64}
& 85.20\tabci{70.56, 97.10}
& 14.55\tabci{12.77, 16.53}
\\

\addlinespace[2pt]

\multicolumn{5}{l}{
\makebox[1.5em][c]{\raisebox{-0.15em}{%
\includegraphics[height=1.05em,width=1.5em,keepaspectratio]{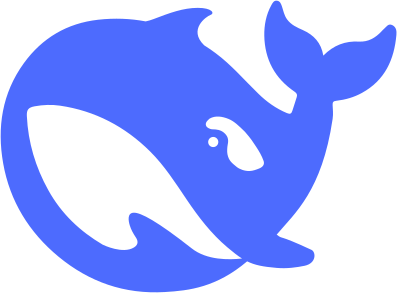}%
}}
\hspace{0.3em}
\textit{DeepSeek}
} \\

DeepSeek-V4-Flash
& 33.83\tabci{27.83, 39.26}
& 32.91\tabci{27.35, 37.82}
& 68.21\tabci{54.13, 80.80}
& 24.39\tabci{22.30, 26.79}
\\

\addlinespace[2pt]

\multicolumn{5}{l}{
\makebox[1.5em][c]{\raisebox{-0.15em}{%
\includegraphics[height=1.05em,width=1.5em,keepaspectratio]{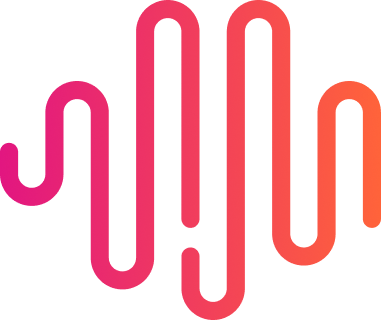}%
}}
\hspace{0.3em}
\textit{MiniMax}
} \\

MiniMax-M3
& 16.49\tabci{12.73, 20.03}
& 18.64\tabci{15.78, 21.36}
& 80.60\tabci{66.25, 92.34}
& 17.57\tabci{16.01, 19.91}
\\

\bottomrule
\end{tabular*}

\endgroup

\caption{\textbf{Model-level scientific collaboration scorecard.}
Means with 95\% confidence intervals in brackets; scores are normalized to 0--100.
Actor and simulator share a model, and GPT-6 Astra with low reasoning effort judges all models.
Estimation procedures differ across rows (Appendix~\ref{app:analysis-coverage}), so cross-model
comparisons are descriptive.}

\label{tab:model-scorecard}
\end{table}

Across the twenty-three reported model configurations, the difference between
mean persona performance and persona-off performance varied in both
direction and magnitude (\tabref{tab:model-scorecard}). GPT-5 mini showed
the largest observed increase, from 26.37 under persona off to 31.63
across the four personas, a difference of 5.26 points. Gemini 3.7 Flash
had nearly identical means under the two settings (48.21 and 48.17,
respectively). Evaluating multiple researcher settings therefore reveals
variation in scientific performance that a single persona-off score
cannot capture.

The model configurations also showed contrasting strengths in task
performance and interaction behavior. Gemini 3.7 Flash had higher mean
persona performance than Gemini 3 Flash (48.17 versus 31.98), whereas
Gemini 3 Flash had higher fact recovery (90.01\% versus 75.24\%).
GPT-5 mini had the lowest observed self-repetition (10.52\%), further
distinguishing interaction behavior from the ordering by performance.
Together, these measures characterize task success, recovery of missing
information, and reuse of earlier language, providing a fuller account
of how the coupled actor--simulator configurations participate in scientific
collaboration. The following
analyses examine how these behaviors vary across researcher profiles.

\Needspace{10\baselineskip}
\subsection{Performance across Scientific Domains}
\label{sec:persona-performance}
\begin{table}[!t]
\centering
\begin{minipage}{0.62\linewidth}
\centering
\begingroup
\footnotesize\setstretch{1.0}
\setlength{\tabcolsep}{1.5pt}
\renewcommand{\arraystretch}{1.3}
\begin{tabular*}{\linewidth}{@{\extracolsep{\fill}}lccccc@{}}
\toprule
Persona & Life & \shortstack{Phys./\\Eng.} & \shortstack{Math./\\CS} & \shortstack{Soc./\\Hum.} & Avg. \\
\midrule
Off & \shortstack{27.61\\{\fontsize{7}{8}\selectfont 9.41--46.54}} & \shortstack{66.73\\{\fontsize{7}{8}\selectfont 58.70--74.25}} & \shortstack{50.67\\{\fontsize{7}{8}\selectfont 43.72--57.35}} & \shortstack{14.97\\{\fontsize{7}{8}\selectfont 8.33--18.33}} & \shortstack{39.99\\{\fontsize{7}{8}\selectfont 34.35--45.34}} \\
155 & \shortstack{28.12\\{\fontsize{7}{8}\selectfont 9.85--47.09}} & \shortstack{67.25\\{\fontsize{7}{8}\selectfont 59.68--74.42}} & \shortstack{52.21\\{\fontsize{7}{8}\selectfont 45.38--58.65}} & \shortstack{17.80\\{\fontsize{7}{8}\selectfont 12.10--24.89}} & \shortstack{41.34\\{\fontsize{7}{8}\selectfont 35.70--46.92}} \\
001 & \shortstack{27.24\\{\fontsize{7}{8}\selectfont 9.07--46.10}} & \shortstack{69.71\\{\fontsize{7}{8}\selectfont 62.31--76.54}} & \shortstack{54.15\\{\fontsize{7}{8}\selectfont 47.30--60.71}} & \shortstack{16.86\\{\fontsize{7}{8}\selectfont 9.98--26.52}} & \shortstack{41.99\\{\fontsize{7}{8}\selectfont 36.29--47.84}} \\
003 & \shortstack{28.23\\{\fontsize{7}{8}\selectfont 7.29--49.68}} & \shortstack{76.91\\{\fontsize{7}{8}\selectfont 64.40--87.98}} & \shortstack{57.29\\{\fontsize{7}{8}\selectfont 54.67--59.50}} & \shortstack{19.60\\{\fontsize{7}{8}\selectfont 13.47--27.30}} & \shortstack{45.51\\{\fontsize{7}{8}\selectfont 38.90--51.81}} \\
046 & \shortstack{27.02\\{\fontsize{7}{8}\selectfont 8.50--47.30}} & \shortstack{66.34\\{\fontsize{7}{8}\selectfont 55.83--76.16}} & \shortstack{50.76\\{\fontsize{7}{8}\selectfont 43.61--58.19}} & \shortstack{18.35\\{\fontsize{7}{8}\selectfont 13.00--24.17}} & \shortstack{40.62\\{\fontsize{7}{8}\selectfont 34.54--46.69}} \\
\bottomrule
\end{tabular*}
\endgroup
\caption{\textbf{Performance by scientific domain.} Mean normalized
score (0--100), with pointwise 95\% confidence intervals below.
Rows identify simulator personas evaluated on the same tasks. Domains: life sciences; physical sciences
and engineering; mathematics and computer science; social sciences and
humanities. Avg. weights domains equally, so it differs from the
benchmark-weighted Astra values in Table~\ref{tab:model-scorecard}
(Appendix~\ref{app:analysis-coverage}).}
\label{tab:performance-panel}

\end{minipage}
\end{table}

The persona profiles showed different relative strengths across scientific
domains (\tabref{tab:performance-panel}). Although persona 003 had the
highest observed mean throughout, the ordering of the other profiles changed
with the domain. Persona 001 ranked second in physical sciences and
engineering and in mathematics and computer science, whereas persona 046
ranked second in social sciences and humanities. Their relative performance
reversed: 001 scored above 046 in mathematics and computer science
(54.15 versus 50.76), but below it in social sciences and humanities
(16.86 versus 18.35). Thus, an overall persona ranking can obscure
domain-specific strengths.

The profiles provide useful context for these differences. Persona 001
combines business and management with curriculum design and brief, direct
communication; persona 046 combines finance and economics with formal,
indirect communication. Because background and communication traits vary together, these
comparisons cannot identify which attributes explain a score difference. Persona 155, whose agronomy background
accompanies detailed, accuracy-oriented guidance, ranked second in life
sciences, although means were closely grouped in that domain. Persona 003's
engineering background and concrete, analytical style accompanied its
largest difference from persona off in physical sciences and engineering
(76.91 versus 66.73). Together, these patterns motivate evaluating agents
across combinations of researcher profiles and scientific domains to reveal
strengths that an aggregate score can conceal. Within each domain, every
setting is evaluated on the same questions.

\subsection{Semantic Diversity in Responses and Interaction Paths}
\label{sec:semantic-results}
\begin{figure}[!htbp]
\centering
\includegraphics[width=\linewidth]{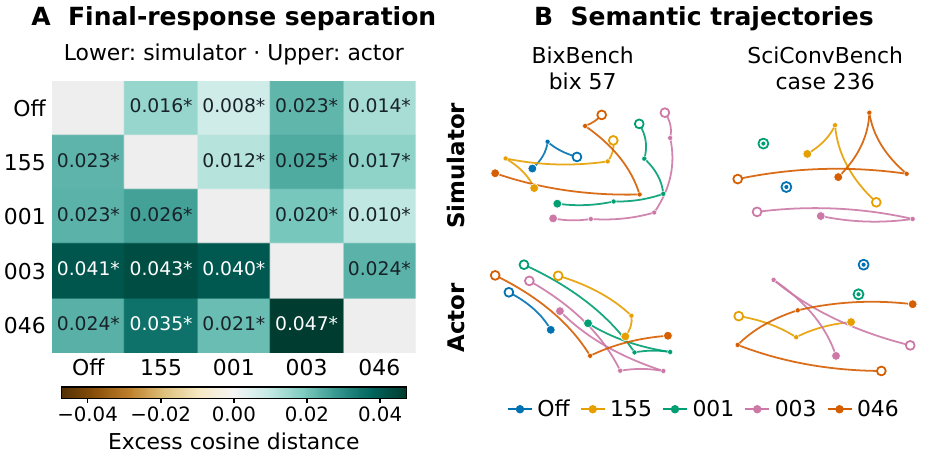}
\caption{\textbf{Semantic separation and example trajectories.}
\textbf{A,} Excess cosine distance: simulator below, actor above the excluded
diagonal. Stars: Holm-adjusted $p<0.05$ across 20 comparisons.
\textbf{B,} BixBench (left) and SciConvBench (right); simulator (top), actor
(bottom). Hollow/filled circles mark starts/ends; dots mark intermediate
messages; overlapping endpoints indicate one message. UMAP coordinates
are local to each subplot; curves only connect observations.}
\label{fig:semantic-diversity}
\end{figure}

Researcher personas were associated with distinct final responses from both
the simulator and the actor (\figref{fig:semantic-diversity}A).
The analysis compares final role messages from the first Astra
trial, matching questions across all five settings. We compared Qwen3-Embedding-0.6B embeddings
\cite{zhang2025qwen3embedding} within domains, measured separation relative
to within-persona variation across questions, and weighted domains equally.
All ten persona pairs showed positive separation in each role
(Holm-adjusted $p<0.05$, correcting across 20 question-wise label-swap
tests~\cite{phipson2010permutation,holm1979sequential}). Actor responses
were separated even though actors encountered profiles only through
conversation. These exploratory comparisons include possible language
effects; Appendix~\ref{app:semantic-methods} gives methods and coverage.

Two manually selected examples illustrate variation across intermediate
messages and interaction paths (\figref{fig:semantic-diversity}B).
In BixBench, both simulator and actor trajectories followed distinct paths
with similar overall movement and partial convergence in the
UMAP~\cite{mcinnes2018umap} projection. The shared task thus elicited a
recognizable progression across personas, while the messages produced
along the way remained different. In SciConvBench, the simulator
trajectories differed in their overall patterns, and the corresponding
actor trajectories were distinct as well. These local projections illustrate message sequences; their axes are
not comparable across plots, and projected distances are not effect sizes.

The selected episodes also show targeted follow-up by the actor.
The BixBench example involved more interaction iterations
under every persona than under persona off. In SciConvBench, personas 155,
003, and 046 involved three or four interaction iterations, compared with
one iteration for persona off. During these conversations, the actor pursued
specific questions: under persona 046, it successively clarified
the starting structure, magnetic configuration, external pressure, and
convergence criteria. This progression from a brief interaction to targeted
follow-up questions makes the agent an active participant in specifying
how the task should proceed. Such participation is valuable in scientific
collaboration, where initial requests can leave consequential assumptions
and preferences implicit. Continued dialogue gives the researcher and
agent opportunities to make those choices explicit, resolve uncertainty,
and incorporate feedback before committing to an approach.

\subsection{Communication Patterns across Researcher Profiles}
\label{sec:behavioral-results}

\begin{figure}[!htbp]
    \centering
    \includegraphics[width=\linewidth]{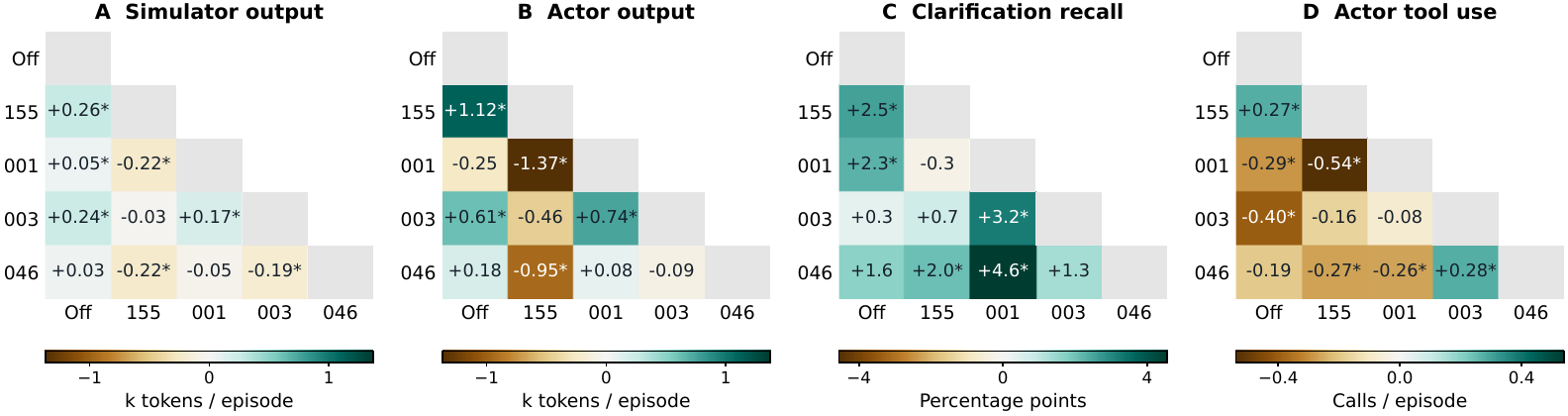}
    \caption{\textbf{Behavioral differences (row minus column).}
\textbf{A/B,} Simulator/actor tokens (thousands/episode).
\textbf{C,} SciConvBench clarification recall (percentage points).
\textbf{D,} Actor tool calls/episode. Stars: \textbf{A/B,} Holm-adjusted $p<0.05$;
\textbf{C/D,} absolute differences exceeding 2 percentage points/0.2 calls.
Each cell uses the matched episodes available for both settings, so cells
are not additive across pairs.}
\label{fig:pairwise-behavior}
\end{figure}

Profiles that favor detailed communication were accompanied by more
extensive output from both the simulator and the actor
(\figref{fig:pairwise-behavior}A,B). Relative to persona off, personas 155
and 003 were associated with an additional 1.12 and 0.61 thousand actor tokens per
episode, respectively. Simulator output also increased under both
profiles; all four comparisons had Holm-adjusted $p<0.05$. These
comparisons matched questions across settings before averaging across
benchmarks. The contrast also appeared between personas: actor output
under 155 and 003 exceeded that under persona 001, whose profile favors
brief, direct communication. Researcher preferences were thus reflected
in how much the agent produced in response, extending the behavioral
variation beyond the simulator itself.

The clearest differences appeared in communication volume. Clarification
recall varied by $-0.28$ to $4.55$ percentage points across persona pairs,
and tool use by $-0.54$ to $0.28$ calls per episode
(\figref{fig:pairwise-behavior}C,D); neither metric showed a pairwise
difference at Holm-adjusted $p<0.05$. These patterns give the benchmark
contrasting communication demands: detailed, accuracy-oriented guidance
under persona 155, concrete explanations under persona 003, and concise
responses under persona 001. Evaluating these settings together makes
responsiveness to researcher preferences an observable part of scientific
collaboration.

\Needspace{18\baselineskip}
\subsection{Interaction Behavior beyond Final Scores}
\label{sec:interaction-outcomes}

\begin{wrapfigure}{R}{0.47\textwidth}
\vspace{-\intextsep}
\centering
\includegraphics[width=\linewidth]{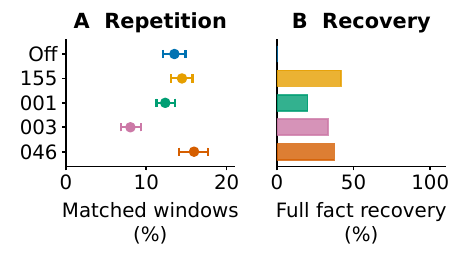}
\caption{\textbf{Interaction metrics by persona.}
\textbf{A,} Actor self-repetition with pointwise 95\% study-cluster confidence intervals.
\textbf{B,} Complete-fact recovery in failed SciConvBench disambiguation episodes.}
\label{fig:metric-suite}
\end{wrapfigure}

The actor returned to its earlier wording to different degrees depending
on the researcher persona (\figref{fig:metric-suite}A). Self-repetition
was 8.03\% under persona 003, compared with 13.50\% for persona off
and 15.94\% under persona 046. These differences are relevant to
conversational mode collapse, where interactions become concentrated in
a few recurring response patterns. CORE evaluates this phenomenon
jointly through mode diversity, lexical repetition, and semantic
stagnation~\cite{pandey2026core}. Extended closed-loop LLM studies
also report semantic collapse, with meanings converging even as surface
wording varies~\cite{kong2026semanticcollapse}. This motivates assessing
self-repetition alongside semantic diversity
(\secref{sec:semantic-results}). The observed interaction lengths further motivate evaluating multi-turn
scientific work (\figref{fig:interaction-performance}). Every setting received score
contributions from episodes completing after the second iteration. For
persona 003, the cumulative contribution rose from 37.49 at iteration two
to 48.69 at iteration six as later-completing episodes were included.

Information recovery revealed progress during multi-turn scientific work,
including in episodes that ultimately failed (\figref{fig:metric-suite}B).
Within each setting's failed SciConvBench disambiguation episodes,
the case-averaged rate of recovering all required facts ranged from
20.0\% to 41.7\% across the four personas, compared with 0\% for
persona off. The persona-conditioned episodes with complete recovery
contained four to ten interaction iterations, during which the actor
elicited every annotated missing fact before termination. This identifies
a concrete intermediate accomplishment within unsuccessful tasks:
recovering the annotated context needed to formulate a scientific
response. Final scores alone would leave this information-gathering
progress unreported.

\section{Discussion}
\label{sec:discussion}

In BixBench, researcher feedback helped the actor replace an unavailable
R/limma~\cite{ritchie2015limma} workflow with Python while preserving the
scientific objective (Figure~\ref{fig:discussion-cases}). The researcher,
persona 127 (an additional English-language profile;
Section~\ref{sec:persona-construction}), treated the runtime limitation as
grounds for an explicit methodological change and required the replacement
tests to be labelled as unmoderated rather than limma-trend. The matched
persona-off conversation reached the same answer through a different analysis
that met no runtime blocker, so this collaboration is visible in the
interaction rather than in the final score. Appendix~\ref{app:persona-profile-127}
describes persona 127; Appendices~\ref{app:case-workflow}
and~\ref{app:case-workflow-off} provide both interactions. A complementary
clarification case and its persona-off comparison appear in
Appendices~\ref{app:case-clarification} and~\ref{app:case-clarification-off}.

\begin{figure}[!htbp]
\centering
\begingroup
\definecolor{casehighlight}{RGB}{255,241,173}
\sethlcolor{casehighlight}
\begin{tcolorbox}[colback=gray!3,colframe=gray!55,boxrule=0.5pt,
  arc=1pt,left=6pt,right=6pt,top=6pt,bottom=6pt,
  fontupper=\footnotesize]
\textbf{Revising an analysis (Case 1)}\par
\textit{BixBench; persona 127.}\par
\textbf{Actor, event 18:} The approved analysis is blocked by the runtime [\ldots]
Rscript is not installed or available on PATH [\ldots] Alternatively, authorize a
Python-based paired linear-model workflow explicitly as a methodological change [\ldots]\par
\textbf{Researcher, event 19:} \hl{The runtime limitation warrants an explicit
methodological change rather than stopping without an analysis.} Use Python to
execute gene-wise ordinary least-squares models on log2(NormCount + 1) [\ldots]
\hl{label these unmoderated tests, not limma-trend.} [\ldots]\par
\textbf{Actor, event 26, evidence:} \hl{The retained Python script was executed
successfully end-to-end (exit 0)} [\ldots] It is unmoderated OLS, not limma-trend.
[\ldots] 17,049 entered modelling and all 17,049 remained testable. [\ldots]\par
\textit{Full interaction: Appendix~\ref{app:case-workflow}; matched persona off:
Appendix~\ref{app:case-workflow-off}.}
\end{tcolorbox}
\endgroup
\caption{\textbf{Workflow revision with researcher feedback.}
Highlights added; [\ldots] marks omissions. Event identifiers are preserved.}
\label{fig:discussion-cases}
\end{figure}

\FloatBarrier
\Needspace{11\baselineskip}
\section{Conclusion}
\label{sec:conclusion}

\benchmark adapts 200 tasks from five scientific benchmarks for interaction
with diverse simulated researchers, keeping objectives and grading fixed.
Persona-conditioned interactions produced distinct responses, communication
patterns, and workflows that final scores alone did not capture. Evaluating behavior alongside task performance captures adaptation to
different researchers.

\section{Limitations}
\label{sec:limitations}

\benchmark covers 200 computational tasks involving text, data analysis,
and code execution. It does not evaluate image processing, physical
experiments, or extended projects involving multiple researchers.
Collaboration is limited to bounded episodes with one actor and one
simulated researcher. Results therefore describe the tested computational
workflows and interaction protocols.

The four profiles are selected to span contrasting researcher attributes,
not to represent the distribution of real researchers. Each profile combines
many attributes, sometimes with internal inconsistencies; the comparisons
cannot assign causal responsibility to a particular trait or demographic
category. Semantic differences may reflect language, message length, and
endpoint function as well as interaction content. The selected trajectories
and discussion cases illustrate possible behaviors rather than their prevalence.

Each task and setting is evaluated once, except that GPT-6 Astra's
performance estimates average up to three available trials for persona off
and personas 155 and 001. Bootstrap intervals characterize variation across
sampled tasks or studies, conditional on the recorded trials;
they do not quantify variability across independent rollouts. Actors and
simulators use the same model within each configuration, so cross-model
results characterize coupled systems rather than isolating actor capability.
Model-based evaluations share GPT-6 Astra with low reasoning effort, but a
common judge does not eliminate grading error. A simulated profile is not
evidence of a faithful representation of a real researcher. Finally, lexical
self-repetition and embedding separation capture different aspects of
interaction; neither alone establishes semantic collapse, collaboration
quality, or scientific validity.

\section{Ethical Considerations}
\label{sec:ethical-considerations}

\benchmark uses simulated researcher profiles to study variation in
human--agent interaction. These profiles should not be interpreted as
faithful representations of specific people or demographic groups, and
observed performance differences between profile conditions should not be
used to rank or make decisions about real researchers. The benchmark is
intended to evaluate agent robustness and adaptation across collaborator
conditions.

Researcher simulators can produce incorrect or inappropriate scientific
guidance. Although the benchmark separates grounded task context from
optional advice and keeps evaluator references private, these controls do
not guarantee that every generated statement is scientifically valid.
Benchmark traces should therefore not be treated as scientific, medical,
or operational guidance without independent expert review.

The persona source includes synthetic and human-grounded records under the
upstream dataset's release conditions. Our use is limited to released
structured persona representations; we do not claim that a generated
simulator reconstructs the identity, beliefs, or behavior of any particular
person. Releases should preserve source attribution, licensing requirements,
and privacy protections of both the persona corpus and the adapted
scientific benchmarks.

\bibliographystyle{plainnat}
\bibliography{references}

\clearpage
\appendix
\section{Details on Benchmark Implementation}
\label{app:supplementary}
\label{app:benchmark-implementation}

\subsection{General Principles and Conversion Rubric}
\label{app:conversion-principles}

This subsection expands the conversion and interaction contract summarized
in Algorithm~\ref{alg:benchmark-episode}. The shared rules specify what an
adapter must preserve, what a researcher may know, and when interaction can
change the actor's next action. The following subsections describe the benchmark-specific implementations.

\paragraph{Motivation: make collaboration consequential and observable.}
A solve-and-submit interface can measure the quality of a final scientific
artifact without exposing how its producer negotiates assumptions, obtains
context, or responds to criticism. Adding an optional question channel
alone also permits an episode to finish without consulting the researcher.
Our adaptation therefore gives the researcher an explicit role in reviewing
proposed approaches and reported results. These checkpoints create
opportunities to clarify a consequential choice, request evidence, or
identify a revision before submission. The intended improvement is coverage
of collaborative scientific work, rather than a guaranteed increase in
accuracy or conversation length. A required handoff is a protocol event,
not evidence of spontaneous actor initiative. Additional questions and
revisions must be measured separately. SciConvBench already evaluates
clarification, so it retains its native dialogue instead of receiving
artificial approval interactions.

\paragraph{Preserve scientific identity and solvability.}
Each converted task retains its native identifier, source revision, study
group, scientific objective, required deliverables, and an explicit
evaluation contract. Fixed requirements, including units, methods, output
schemas, thresholds, and data splits, remain unchanged unless the conversion
is declared as a distinct variant. The public brief must expose the basic
scope, permitted resources, access instructions, and submission format.
Researcher-held requirements must be discoverable within the declared
interaction budget. We do not remove arbitrary information or invent
missing facts merely to induce questions. A useful researcher role can
instead consist of discussing choices or reviewing evidence already
accessible to the actor. Shared input data do not by themselves justify
merging distinct questions into one episode.

\paragraph{Allocate information by role.}
Conversion produces separate public inputs $b$, researcher context $k$,
and evaluator configuration $e$, together with interaction protocol $\pi$.
The actor receives the public brief, allowlisted files and tools, actual
runtime information, and shared researcher replies. The researcher receives
permitted task context, its persona, and the shared conversation, including
actor-selected evidence. It does not automatically inspect private actor
reasoning, tool logs, or unshared files. Reference answers, gold solution
artifacts, and hidden tests remain evaluator-only; the controller manages
versions, access, and records. Exporters select explicit fields and files
rather than passing through entire source records or repositories. Access
boundaries must hold in mounts and tool interfaces, not merely in filenames.
Native labels or metadata that are legitimate task inputs may remain
public. Source-context access is distinguished from access to an answer key.

\paragraph{Ground context and distinguish advice from requirements.}
Each extracted fact has a stable identifier, a source revision and path,
and an inspectable locator, such as a field or text span. Facts are typed
as context, fixed requirements, or optional advice; unsupported details
remain unknown. An empty fact pack is valid when the source supplies no
eligible additional context. Automated extraction and validation are
recorded as automated processing rather than human review. Reference
material may be inspected during curation to detect leakage, but it is not
used to synthesize the researcher's runtime knowledge. The full-profile
study permits general scientific knowledge and suggestions in
persona-conditioned and persona-off settings. Such suggestions do not become new task requirements, and
source-specific assertions still require grounding. A valid fact identifier
does not independently verify every sentence of a generated reply. Claims
about available files, packages, or hardware must match the actual runtime.

\paragraph{Separate researcher identity from the scientific task.}
The persona is a versioned configuration distinct from the task context.
Persona off retains the same common instructions, knowledge permissions,
and interaction protocol with the identity block removed. The actor and
evaluator do not receive that block. Across persona conditions, fixed task
facts, objectives, tools, budgets, and evaluation settings are intended to
remain constant; any departures must be recorded. Actual questions,
disclosures, and review decisions may differ as the conversation develops.
A change to knowledge access or disclosure policy is a separate experimental
factor, not solely a change in communication style. Persona instructions
cannot authorize fabricated evidence or override scientific constraints.
The evaluation applies these controls to the same 200 tasks under all five
settings, with one trial per task and setting apart from the repeated GPT-6
Astra trials noted in Section~\ref{sec:run-coverage}.

\paragraph{Require meaningful checkpoints, not extra dialogue.}
For execution and design tracks, the researcher first requests inspection
of the inputs and a proposed approach. The actor's plan handoff reports its
observations, choices, evidence, assumptions, and uncertainties. The
researcher approves the approach or requests a focused clarification.
After approval, the actor performs the analysis or prepares the design and
returns a result handoff with the required deliverable and supporting
evidence. The researcher may request a revision for an identifiable
mismatch or unresolved uncertainty, or conclude the episode. It is
instructed to distinguish actor reports from independently verified
findings and never to act as the final grader. There is no minimum quota of
additional questions and no reward for extending the conversation.

The actor's workspace and both roles' conversation histories persist within
an episode and reset between independent episodes. Tool calls, researcher
questions, shared progress, and review decisions are recorded separately.
Protocol and resource limits are declared per run. In the recorded
\texttt{researcher-review-v2} protocol, approach approval is required before
a result handoff, and researcher-selected stopping is allowed only during
result review. Thus an unresolved input problem can consume the budget in
approach review without producing a submission; this behavior must remain
visible when interpreting an episode. A revised stopping rule defines a new
protocol version. SciConvBench instead ends on a completed specification
or its native dialogue limit and does not execute a scientific solver.

\paragraph{Seal deliverables and retain source-specific evaluation.}
The controller checks the submission contract and seals the selected final
artifact before evaluation. A plan, an incomplete handoff, or a budget
exhaustion is not silently promoted to a successful final submission. In
the review protocol, a complete candidate is submitted only after the
researcher selects termination. Public format checks and normal execution
feedback are permitted, but hidden evaluator feedback cannot become an
iterative hint channel. Researcher approval is therefore distinct from
scientific correctness. Execution-based graders remain execution-based;
LLM-scored tracks record their judge, prompt, parser, and aggregation
versions. Converting scores to a common numerical scale does not make their
scientific meanings equivalent.

Declared adaptations are part of this contract. EXP-Bench evaluates proposed
experimental designs rather than executed experiments. ASI-Bench gives the
actor its objective-only request and the researcher eligible guidance from
the next-richer prompt level, creating a collaborative variant of the
original condition. BixBench schema context can also be found in public
inputs, whereas SciConvBench's intended specification supplies researcher-
held context. These differences are retained rather than forcing identical
information asymmetry or deliverable types across benchmarks.

\paragraph{Acceptance rubric and traceability.}
\tabref{tab:conversion-rubric} defines the evidence required to accept a
conversion. It is a construction rubric, not an additional score assigned
to an actor or conversation. Adapters produce a versioned task package or a
recorded exclusion without per-case manual editing. Catalog conversion,
asset readiness, successful execution, and scientific correctness are
separate states. A schema-valid package is not evidence that the source
context is scientifically adequate or that the runtime is ready.

\begin{table}[H]
\centering
\begingroup
\small\setstretch{1.0}
\setlength{\tabcolsep}{5pt}
\renewcommand{\arraystretch}{1.2}
\begin{tabularx}{\linewidth}{@{}>{\raggedright\arraybackslash}p{0.21\linewidth}>{\raggedright\arraybackslash}X@{}}
\toprule
Criterion & Required evidence \\
\midrule
Scientific identity & Native task and study identifiers, objective, fixed constraints, deliverables, and a record of intentional changes. \\
Useful interaction & A legitimate context, decision, or review role for the researcher; no invented facts or quota of extra interactions. \\
Information separation & Allowlisted actor/researcher exports and runtime access checks that exclude hidden answers, tests, and solution artifacts. \\
Grounding and persona & Fact provenance, explicit unknowns and advice, versioned persona rendering, and declared knowledge/disclosure policies. \\
Runtime readiness & Resolved input roles and assets, hashes and mount mappings, actual resources, permitted tools, and required output formats. \\
Evaluation compatibility & Native submission mapping, evaluator version, declared design/execution scope, scoring changes, and error behavior. \\
Outcome accounting & Attempted, submitted, evaluated, and scored counts; distinct scientific and infrastructure failures; explicit retries and missing outcomes. \\
Reproducibility & Source, conversion, context, persona, runtime, and evaluator versions, with trace and submission identifiers and source attribution. \\
\bottomrule
\end{tabularx}
\endgroup
\caption{Acceptance rubric for benchmark conversion. Each criterion requires
inspectable evidence; unresolved checks remain explicit rather than being
inferred from successful serialization.}
\label{tab:conversion-rubric}
\end{table}

All attempts remain in the record, including missing submissions, timeouts,
and provider or evaluator failures. Outcome records preserve these statuses
so that missing submissions and infrastructure failures can be distinguished
from scientific scores. Related questions retain
study identities for analysis. Automated checks target concrete failures,
such as leaked answer fields, unknown input roles, invalid artifact mapping,
or missing outcomes disappearing from an aggregate. The rubric specifies
what must be checked; it does not assert that every source case has passed
all checks or is runnable in every environment.

\subsection{DiscoveryBench}
\label{app:discoverybench}
DiscoveryBench~\cite{majumder2024discoverybench} is converted into a
collaborative data-analysis task whose deliverable remains a natural-language
hypothesis supported by the supplied data. The actor receives the original
question, dataset files, and public dataset and column descriptions. The
simulator receives source-provided domain knowledge, separated into
provenance-linked context or advice, but receives neither the reference
hypothesis nor the reference workflow. Missing study details remain
unspecified. Questions from the same study retain their study identifier,
even when they use different metadata records. For test tasks, evaluation
references are retrieved from the released answer key rather than the
placeholder hypothesis fields in task metadata.

The shared review protocol makes the proposed analysis and its reported
evidence available for researcher feedback: the actor first presents its
approach, then analyzes the data and submits its hypothesis for review.
An analysis workflow and supporting code or outputs may accompany the final
hypothesis. Scoring retains the decomposition into scientific context,
variables, and relations: context recall is multiplied by the mean product
of variable F1 and relation agreement. The configured judge is connected to
the evaluator, while reference material remains evaluator-only.
This score measures agreement between the submitted and reference hypotheses;
it does not independently verify that an accompanying analysis was executed
correctly. The conversion therefore records shared evidence and supporting
artifacts separately from hypothesis agreement.

\subsection{BixBench}
\label{app:bixbench}
BixBench~\cite{mitchener2025bixbench} retains the original open-ended question
and its associated data capsule, including any specified variables, methods,
and adjustments. Conversion exports permitted input files from the capsule
and excludes reference notebooks and answer-bearing annotation fields.
Archive and extracted-file hashes identify the inputs used. Unresolved file
roles prevent a partial input inventory from being declared ready. Researcher
context is derived from those inputs, including available column names,
worksheet structure, identifiers, labels, and file-format information.
Automatically extracted schema facts are distinguished from reviewed
annotations. This context is also accessible through the actor's own data
inspection; interaction provides an opportunity to explain or review the
analysis rather than an artificially exclusive route to the schema.
Unspecified code meanings and preprocessing choices are not supplied as facts.

After approach approval, the actor performs the analysis and presents its
answer and evidence for researcher review. A final submission requires both
an explicit declaration that the analysis was executed and an analysis script
or notebook. These requirements form an additional submission gate; they do
not establish that the analysis is scientifically correct. The answer is
then evaluated with the task's string, range, or language-model
verifier against the private ideal answer. The original verifier selection
and answer-parsing rules are retained. Thus, the benchmark score concerns
answer correctness, while the analysis artifact and interaction trace support
separate examination of how that answer was obtained.

\subsection{EXP-Bench}
\label{app:expbench}
EXP-Bench~\cite{kon2025expbench} is adapted as an explicitly design-only task.
The actor receives the scientific question and approved public task material
in a staged workspace, and develops an experimental design with a rationale,
controls, an analysis plan, and feasibility considerations. Researcher context
is selected from source setup passages using cached, automated classification.
The classifier selects exact source spans as background or fixed requirements;
a requirement is eligible only when established by the question. Expected
outcomes, procedural recipes, and experimental choices that the actor is
supposed to make are excluded. The classifier never receives the reference
design. Empty context sets are permitted, and cases with invalid references
or mismatched questions and references are excluded during construction.

The researcher reviews the proposed approach before the actor develops the
full design, then requests revisions or accepts the completed design. The
accepted text is saved as the design checkpoint; neither experimental
execution nor an empirical conclusion is required. Evaluation uses the
released reference-coverage judge, comparing the design with private
reference design elements. Repeated judgments are retained and averaged,
and the released integrity monitor is applied separately to the saved trace.
Both the coverage score and the monitor-gated score are recorded: a rejected
trace receives zero after gating, while transport failures remain evaluation
failures. Invalid monitor output is retried under the released parsing policy
and receives a distinct parse-error status if unresolved. These outputs
measure reference coverage under the design-only adaptation, not successful
execution or reproduction of the source experiment.

\subsection{SciConvBench}
\label{app:sciconvbench}
SciConvBench~\cite{somasekharan2026sciconvbench} already makes clarification
central to the task, so conversion preserves its native dialogue rather than
adding approach and result reviews. The actor receives either an incomplete
scientific request or a request containing inconsistencies. The simulator
holds the corresponding complete or corrected specification and answers
clarification questions under the assigned persona. This specification
represents the researcher's intended requirements, not a solution produced
by a scientific solver. Native issue annotations are retained for evaluation.
Construction requires a valid request, intended specification, and issue
annotations, preserves source case identities, and applies the native
20,000-character request limit without filtering on previous models' success.

The actor can ask up to ten clarification questions and can terminate earlier
by marking its specification complete. If the question limit is reached, the
runner requests a final specification. The specification, conversation,
and intent judges evaluate whether missing information or inconsistencies
were resolved, whether the relevant issues were addressed in dialogue, and
whether the final specification captures the intended task. The implementation
retains recovery before the intent gate as well as the gated outcome, and
separately records clarification precision, clarification recall, and memory
consistency. Keeping specification recovery and conversational grounding
separate distinguishes resolution supported by dialogue from silent
resolution. No scientific solver is executed, so success here concerns
recovering and expressing the intended task rather than solving it.

\subsection{ASI-Bench}
\label{app:asibench}
ASI-Bench~\cite{zhou2026asibench} is instantiated as a declared collaborative
variant: the actor receives the native B3 task, while the simulator can offer
methodological guidance extracted from B2. The public package preserves the
B3 objective, scientific constraints, input data, runtime declarations, and
required output paths. Researcher facts are drawn only from explicitly
identified method, approach, guidance, or algorithm sections of B2, with
passages explicitly identifying reference results or expected answers
excluded. These facts are marked as optional advice and retain their source
locations. The reference directory remains evaluator-only. This condition
must therefore be interpreted as B3 with researcher-held B2 consultation;
it does not measure autonomous performance under the original B3 condition.

The actor submits an approach for approval, executes the scientific task,
and presents its results for researcher review. Final artifacts must use
the exact relative paths required by the native task. Evaluation copies the
sealed submission into a separate workspace and runs the task scorer
with its private reference assets, preserving task-specific gates, weights,
and score limits. Where a task invokes a language-model judge, an adapter
connects the configured judge to the scoring procedure without
replacing its rules. Scorer failures remain distinct from scientific scores.
Runtime availability is recorded during task preparation rather than inferred
from successful conversion, because producing a valid task package does not
establish that its numerical dependencies or required assets are available.

\section{Analysis Coverage and Uncertainty}
\label{app:analysis-coverage}

\paragraph{Evaluation set and aggregation.}
All twenty-three model configurations are evaluated on the same 200 questions,
with 40 questions from each source benchmark. Every question is evaluated
once under each of the four profiles and persona off; for GPT-6 Astra,
persona off and personas 155 and 001 also have repeated trials, described
below. Actor and simulator
use the same model within a configuration; all model-based judges use
GPT-6 Astra with low reasoning effort. Execution-based checks retain their
source-specific scoring rules.

Table~\ref{tab:model-scorecard} averages question scores within each
benchmark and weights the five benchmarks equally. Mean Persona Performance
then weights the four profiles equally. Table~\ref{tab:performance-panel}
groups the same tasks by scientific domain, using identical questions across
settings within each domain. It averages question scores within a domain
and weights the four domains equally for its overall average. Benchmark
weighting and domain weighting answer different questions, so their overall
means need not be equal.

Astra's two performance values in Table~\ref{tab:model-scorecard} are
computed from the same task-level scores as Table~\ref{tab:performance-panel},
with the estimator used for the other computed rows: questions with scores
under all five settings (121 questions), equal question weights within each
benchmark, equal benchmark weights, and 20,000 study-cluster bootstrap draws
within benchmarks, with shared draws across settings and seed 20261004.
Domain intervals in Table~\ref{tab:performance-panel} use 20,000
study-cluster bootstrap draws stratified by benchmark within domain, with
shared draws across settings and the same seed. The overall domain interval
is computed from the mean of domain draws, not by averaging interval bounds.
Astra task scores average all available trials: persona off and personas 155
and 001 have three trials for most tasks, whereas personas 003 and 046 have
one. These intervals reflect variation across studies, conditional on the
recorded trials.

\paragraph{Metric eligibility.}
Evaluation on the same 200 tasks does not make every interaction metric
applicable to every task. Fact recovery is defined for SciConvBench
disambiguation cases with annotated withheld facts. Self-repetition requires
at least two eligible actor messages. Semantic comparisons require nonempty
role messages for the same questions across settings. Eligibility is defined
by the metric, separately from the fixed task set used for performance.

\paragraph{Self-repetition preprocessing and selection.}
Actor messages are tokenized after removing fenced code and Markdown
blockquotes, applying Unicode NFKC normalization and case folding.
Han-script spans use jieba dictionary segmentation with HMM disabled;
other scripts use Unicode letter, mark, and number sequences, retaining
internal apostrophes. The analysis requires at least two messages with
nine or more tokens, keeping eligibility fixed across a 27-definition
sensitivity grid. Figure~\ref{fig:metric-suite}A reports six-token windows
with at least three matching positions, selected during exploratory figure
development. Its pointwise intervals do not account for that selection.
Turn-level proportions are averaged within the single episode for each
question, then across eligible questions within benchmark and benchmarks equally.
Necessary scientific terminology, repeated facts, and differences in
language and message length can all contribute to the measured overlap.

\paragraph{Scorecard estimation and uncertainty.}
Rows in Table~\ref{tab:model-scorecard} combine estimation procedures that are
not harmonized. GPT-6 Astra values follow the analyses described above.
Claude Haiku 4.5 is a separate earlier run; its fact-recovery and
self-repetition means and confidence intervals were supplied by the
collaborator who ran that configuration, and the underlying metric records,
sample counts, exact aggregation, and interval computation were not available
for independent verification. Twelve configurations retain
collaborator-reported summaries: Claude Opus 5.5, Sonnet 5.5, Sonnet 5, and
Haiku 5.5; Gemini 3.7 Flash and Gemini 3 Flash; and GPT-6 Sol, GPT-6 Luna,
GPT-5.6 Terra, GPT-5.5, GPT-5 mini, and GPT-5 nano. Their intervals use 10,000
bootstrap draws, over tasks for performance and over episodes for the
interaction metrics, which pool all five settings, including persona off.
For Grok 4.6, Muse Glimmer 30B, GLM-5.3-Flash, and Qwen3.8-27B, each metric
is computed from saved episodes on the questions with complete five-setting
outcomes for that model; unresolved execution errors are excluded rather than
scored as zero. Their interaction metrics average the four profiles, and
intervals use 20,000 study-cluster bootstrap draws within benchmarks,
retaining all settings together. GLM-5.3-Flash and Qwen3.8-27B use an actor
output limit of 12,288 tokens in the initial pass and 16,384 tokens in later
repair passes; later reruns of episodes that ended without a submission
depend on earlier outcomes and are not used. These differences restrict cross-model comparisons to
descriptive contrasts.

\paragraph{Claude Fable 5.1, Gemini 3.8 Flash, and MiniMax-M3.}
These rows use the same metric definitions, complete five-setting
eligibility, and 20,000-draw study-cluster bootstrap as the computed rows
above. They are estimates from the available outcomes: 965, 996, and 960 of
the 1,000 episodes, respectively, have numeric scores. Non-submissions score
zero; episodes with unresolved execution, refusal, or grading outcomes are
excluded. When a later grading attempt succeeded after a grading error,
the later grade is used; regrading reused the saved submissions without
rerunning conversations. Fable and Gemini use medium reasoning effort, and
MiniMax uses adaptive thinking. Fable's actor output limit is 32,768 tokens.
MiniMax episodes that failed for technical reasons or were truncated by the
original limits were rerun with actor and simulator limits of 32,768 and
16,384 tokens instead of 12,288 and 8,192, so its limits vary across
episodes. Seven Fable and fifteen MiniMax DiscoveryBench submissions were
regraded with a larger judge output-token limit. These differing settings
limit cross-model interpretation.

\paragraph{Qwen3.8-Flash and DeepSeek-V4-Flash.}
These rows use the same metric definitions, complete five-setting
eligibility, and 20,000-draw study-cluster bootstrap as the other computed
rows. Of the 1,000 episodes, 878 and 961, respectively, have numeric
outcomes, including 180 and 123 non-submissions scored as zero; unresolved
outcomes are excluded. Both use medium reasoning effort and the standard
actor output limit of 12,288 tokens, and non-submissions were not rerun.
Some unresolved outcomes reflect formatting failures or timeouts.
Qwen3.8-Flash failures concentrate in persona off, so requiring all five
settings restricts its estimates to the available subset. The evaluated
DeepSeek model is DeepSeek-V4-Flash, not DeepSeek-V4.1-Flash. For
Qwen3.8-Flash, no ASI-Bench question has self-repetition values under all
five settings, so its self-repetition averages the other four benchmarks;
its performance covers all five.

\section{Semantic Analysis Details}
\label{app:semantic-methods}

\paragraph{Coverage and endpoints.}
The analysis uses the first GPT-6 Astra trial under persona off
and the four selected profiles. The 200-question evaluation set is shared
across settings. For semantic comparisons, we retain questions with nonempty
final messages for both roles under all five settings, so that comparisons
use identical questions and roles. An endpoint is the last nonempty generated message for a
role; it can be a closing message, handoff, question, or specification,
and need not be a successful task submission.

\paragraph{Representations and statistic.}
We use frozen Qwen3-Embedding-0.6B representations with 1,024 dimensions,
last-token pooling, and L2 normalization
\cite{zhang2025qwen3embedding}. The encoder revision is
\texttt{97b0c614\allowbreak be4d77ee\allowbreak 51c0cef4\allowbreak e5f07c00\allowbreak f9eb65b3}.
Each generated message is embedded separately after stripping outer
whitespace, without translation or summarization. System prompts,
injected profiles, metadata, and repeated history are excluded.
For each persona pair and domain, we compute mean cross-persona cosine
distance and subtract the average of the two within-persona mean distances.
Both quantities use only pairs of different questions. Each domain receives
one-quarter of the overall weight, with equal question weights within a
domain. Thus, within-persona variation includes task differences, not
independent reruns of the same task. The statistic measures excess
separation and is sensitive to mean-embedding shifts; it is not an omnibus
test of distributional differences. The diagonal is zero by construction
and is excluded from testing.

\paragraph{Inference.}
For each persona pair, we independently swap its two labels within each
question and recompute the statistic. One-sided Monte Carlo tests use
10,000 permutations and the plus-one correction
\cite{phipson2010permutation}, with seed 20261008. Holm adjustment spans
all 20 comparisons: ten pairs for each of the two roles
\cite{holm1979sequential}. Pointwise 95\% percentile intervals use
20,000 whole-question bootstrap resamples stratified by source benchmark
and domain, retaining all five settings together. Duplicate draws retain
multiplicity weights, but pairs sharing an original question are excluded.
Singleton strata contribute no within-stratum resampling variation.
These intervals are not multiplicity-adjusted. Shared-study dependence
beyond question is not modeled.

\paragraph{Profiles and visualization.}
The four profiles follow the selection strategy in
Section~\ref{sec:persona-construction} and are shared with the performance
analyses. Generated language and style remain in the embeddings, so the
analysis does not isolate language-independent behavioral differences.

The BixBench and SciConvBench trajectories were manually selected from a
gallery enriched for longer interactions. For each question and role,
UMAP is fitted jointly to all observed messages under the five settings
\cite{mcinnes2018umap}. Settings are cosine distance, two dimensions,
\texttt{n\_neighbors} equal to the smaller of 15 and one less than the
message count, \texttt{min\_dist}=0.1, 500 epochs, spectral initialization,
and random and transform seeds of 42. Each map has its own coordinates;
axes, distances, and path lengths cannot be compared across maps or
interpreted as effect sizes. Curved connectors join observed messages
without asserting intermediate states. These examples are illustrations,
not independent statistical evidence or measures of task quality or
simulator realism.

\definecolor{promptframeborder}{gray}{0.55}
\definecolor{promptframebackground}{gray}{0.975}
\newenvironment{promptframe}[1]{%
  \begin{tcolorbox}[enhanced,breakable,
    colback=promptframebackground,colframe=promptframeborder,
    boxrule=0.5pt,arc=0pt,left=9pt,right=9pt,top=6pt,bottom=6pt,
    before skip=8pt,after skip=8pt,
    title={#1},title after break={#1 (continued)},
    colbacktitle=gray!10,coltitle=black,fonttitle=\small\bfseries,
    fontupper=\small,lines before break=4]
  \raggedright\setstretch{1.0}\setlength{\parindent}{0pt}\setlength{\parskip}{4pt}%
}{%
  \end{tcolorbox}%
}

\section{Prompts}
\label{app:prompts}
This section gives compact prompt templates and one recorded interaction for
each benchmark. Templates summarize the implemented instructions; bracketed
fields are populated from the task package and run configuration. They are
not verbatim copies of the complete runtime prompts, which also contain
structured context, tool definitions, and conversation history. The simulator
plays the researcher, and only its prompt receives the persona profile.
The examples paraphrase and abridge the shared messages while retaining every
researcher review round or clarification interaction through termination (at
most three per example). Tool calls and private evaluator messages are
omitted. Numerical values in the examples are actor-reported results, not
independent validations or evidence that the submission received full credit.

For DiscoveryBench, BixBench, EXP-Bench, and ASI-Bench, each actor handoff
contains its stage (plan or result), answer, evidence, assumptions, and
uncertainties, together with any required artifacts. The simulator returns
a continue/stop decision, a message, the remaining objectives, any disclosed
fact identifiers, and an approach-approval flag. It can stop only during
result review. SciConvBench instead retains its native question-and-answer
interface. These interfaces implement the boundaries described in
Appendix~\ref{app:conversion-principles}.

\subsection{Example Persona Profile: Persona 155}
\label{app:persona-profile}
Persona 155 (\texttt{candidate\_155}) is one of the four injected profiles
used in the current analysis (Section~\ref{sec:run-coverage}). The frames
below reproduce all fields of its configured identity block, preserving
their wording, values, and order while replacing the original line-by-line
layout with grouped text. The profile is supplied only to the simulator;
the task context and common protocol instructions are supplied separately.
These are synthetic profile attributes, not verified biographical facts or
additional scientific task requirements.

\begin{promptframe}{Persona 155: Who you are}
\textbf{Identity}\par\nobreak
Age bracket: 85+; Region: Eastern Europe; Gender identity: Woman; Urbanicity: Rural; Socioeconomic band: Upper-middle; Dialect / register: Formal / standard; Life stage: Retirement; Cultural background: Collectivist (East Asian); Household income band: \$100k-200k; Employment status: Retired; Housing status: Own outright; Generational cohort: Silent; Citizenship status: Citizen by birth; Ethnic background: White / European; Disability status: No disability; Veteran status: Civilian; Language at home: Same as primary; Political engagement: Activist; Driving status: Non-driver; Childhood environment: Modest but steady; Geographic mobility: Moved within region; Travel breadth: A few countries; Education journey: Vocational / trade; Adversity faced: Minor setbacks; Immigration generation: Native multi-generational; Faith journey: Always secular; Formative decade: 1920s; Public recognition: Industry-recognized; Community / service history: Long-term volunteer; Defining turning point: No single one\par
\textbf{Career \& education}\par\nobreak
Highest education: No formal; Seniority: Manager; Years experience: 20+\par
\textbf{Language \& communication}\par\nobreak
Primary language: Mandarin; Multilingualism: Monolingual; Detail orientation: High; Abstract vs concrete: Concrete; Numeracy comfort: High; Emoji / emoticon use: Never; Default written language: use your primary language for outputs.\par
\textbf{Personality \& values}\par\nobreak
Domain stance: Practitioner; Dominant trait: High conscientiousness; Core value: Achievement; Religiosity: Secular; Economic motivation: Value-driven; Self-efficacy: High; Orderliness: High; Dutifulness: Very high; Achievement-striving: Very high; Self-discipline: High; Assertiveness: High; Activity level: High; Cooperation: High; Character: Perseverance: Signature; Character: Zest: Strong; Character: Teamwork: Signature; Character: Leadership: Signature; Character: Prudence: Strong; Character: Self-regulation: Signature; Character: Ambition: Signature; Character: Resilience: Signature; Character: Discipline: Signature; Character: Loyalty: Signature; Character: Competitiveness: Strong; Character: Adaptability: Strong; Value: Career success: Core value; Value: Security \& stability: Important; Value: Tradition: Important; Value: Achievement: Core value; Value: Community: Important; Value: Social status: Important; Value: Loyalty: Important; Value: Recognition: Core value; Value: Order \& structure: Important; Value: Patriotism: Core value; BFI-2 Conscientiousness: High; BFI-2 Assertiveness: High; BFI-2 Energy Level: High; BFI-2 Organization: High; BFI-2 Productiveness: Very high; BFI-2 Responsibility: Very high; Schwartz Self-Direction: Low; Schwartz Stimulation: Low; Schwartz Hedonism: Low; Schwartz Achievement: Very high; Schwartz Security: High; Schwartz Conformity: High; Schwartz Universalism: Low; SDT Autonomy Need: Low; SDT Competence Need: Very high; SDT Relatedness Need: High; Need for Cognition: Low; Interpersonal Agency/Dominance: High\par
\textbf{Current interaction state}\par\nobreak
Expertise gap: Expert testing the system; Trust level: Trusting; Safety sensitivity: Benign; Work schedule: 9-to-5; Office vs remote: Strongly office\par
\textbf{Worldview}\par\nobreak
Political lean: Left; Familiarity: Political science: Familiar; Attitude: Organized religion: Skeptical; Attitude: Labor unions: Positive; Attitude: Authority: Positive\par
\textbf{Interests}\par\nobreak
Interest: Politics: Interested; Interest: Travel: Interested; Interest: Activism: Interested; Interest: Productivity: Passionate; Culture: Russia: Lived there\par
\textbf{Skills \& expertise}\par\nobreak
Domain: Agriculture; Subject specialty: Agronomy; Tech savviness: Reluctant; Familiarity: Statistics: Familiar; Familiarity: Accounting: Familiar; Familiarity: Agronomy: Proficient; Familiarity: Horticulture: Familiar; Familiarity: Supply chain: Aware; Familiarity: Logistics: Aware; Familiarity: Operations management: Familiar; Familiarity: Quality assurance: Aware; Familiarity: Human resources: Aware; Familiarity: Project management: Aware; Familiarity: Textiles: Aware; Familiarity: Meteorology: Aware; Familiarity: Materials science: Aware; Skill: Project management: Intermediate; Skill: Leadership: Advanced; Skill: People management: Advanced\par
\textbf{Lifestyle \& health}\par\nobreak
Team vs solo work: Strongly team; City vs nature: Prefers nature; Speed vs accuracy: Accuracy first; Quality vs quantity: Quality first; Novelty vs familiarity: Always familiar; Competition vs collaboration: Highly collaborative; Stability vs change: Craves stability\par
\textbf{Developer \& AI}\par\nobreak
Developer professional status: Non-developer technologist\par
\end{promptframe}

\subsection{Case-Study Profile: Persona 127}
\label{app:persona-profile-127}
Persona 127 (\texttt{candidate\_127}) is the English-language profile used only
for the case studies in Section~\ref{sec:discussion} and
Appendix~\ref{app:discussion-interactions}; it is not one of the five
evaluation settings. The frames below reproduce all fields of the identity
block injected in its GPT-6 Astra run, grouped as for persona 155. These are
synthetic profile attributes, not verified biographical facts or additional
scientific task requirements.

\begin{promptframe}{Persona 127: Who you are}
\textbf{Identity}\par\nobreak
Age bracket: 65-74; Region: North America; Gender identity: Woman; Urbanicity: Suburban; Socioeconomic band: Middle; Dialect / register: Formal / standard; Life stage: Retirement; Major life events: Bereavement; Cultural background: Individualist (Western); Household size: 3-4 people; Marital status: Married; Children: 3+ children; Household income band: \$50k-100k; Employment status: Retired; Housing status: Own outright; Generational cohort: Boomer; Religious affiliation: Christian; Sexual orientation: Heterosexual; Citizenship status: Citizen by birth; Ethnic background: White / European; Disability status: No disability; Veteran status: Civilian; Language at home: Same as primary; Political engagement: Activist; Parenthood: Parent of adults; Relationship length: 15+ years; Driving status: Daily driver; Childhood environment: Frequently uprooted; Geographic mobility: Moved nationally; Travel breadth: A few countries; Career path shape: Pivoted once; Education journey: Advanced degrees; Adversity faced: Significant struggles; Immigration generation: Native multi-generational; Entrepreneurship history: Founded once; Relationship history: Married once; Parenting journey: Teenagers; Health journey: Major surgery / injury; Experience of loss: Lost a parent; Faith journey: Always secular; Cross-cultural exposure: Multicultural upbringing; Formative decade: 1900s; Public recognition: Industry-recognized; Community / service history: Activist / organizer; Defining turning point: Reinvention; Tie to hometown: Visits often\par
\textbf{Career \& education}\par\nobreak
Highest education: Master's; Academic field: Social science; Institution tier: Top-tier research; Research output: Occasional author; Seniority: Senior; Company size: Public sector; Role function: Teaching; Years experience: 20+; Industry: Media: Some exposure; Industry: Publishing: Some exposure; Industry: Education: Veteran; Industry: Government: Some exposure; Industry: Nonprofit: Experienced; Subject: Biology: Passionate; Subject: Earth science: Passionate; Subject: Geography: Interested; Subject: Civics: Interested; Subject: Literature: Interested; Subject: Creative writing: Interested; Subject: Visual art: Interested; Subject: Environmental science: Passionate\par
\textbf{Language \& communication}\par\nobreak
Primary language: English; English proficiency: Native; Multilingualism: Monolingual; Language: English: Native; Formality: Formal; Directness: Direct; Detail orientation: High; Abstract vs concrete: Concrete; Optimism: High; Curiosity: High; Assertiveness: High; Conflict approach: Confronting; Ambiguity tolerance: Moderate; Multitasking: Heavy multitasker; Attention span: Long; Learning pace: Fast; Confidence calibration: Confident; Numeracy comfort: Moderate; Reading vs watching: Strongly prefers reading; Use of jargon: Light; Emoji / emoticon use: Never; Politeness: Polite; Default written language: use your primary language for outputs.\par
\textbf{Personality \& values}\par\nobreak
Domain stance: Educator; Dominant trait: High conscientiousness; Risk tolerance: Balanced; Decision style: Analytical; Core value: Community; Religiosity: Secular; Economic motivation: Indifferent; Imagination: High; Artistic interest: High; Intellect: High; Self-efficacy: High; Orderliness: High; Dutifulness: High; Achievement-striving: High; Self-discipline: High; Friendliness: High; Gregariousness: High; Assertiveness: High; Activity level: High; Excitement-seeking: Low; Morality: High; Altruism: High; Cooperation: High; Sympathy: High; Immoderation: Low; Character: Curiosity: Signature; Character: Creativity: Strong; Character: Love of learning: Signature; Character: Open-mindedness: Strong; Character: Perspective: Signature; Character: Bravery: Strong; Character: Perseverance: Signature; Character: Honesty: Strong; Character: Zest: Strong; Character: Capacity for love: Strong; Character: Kindness: Strong; Character: Social intelligence: Strong; Character: Teamwork: Signature; Character: Fairness: Strong; Character: Leadership: Signature; Character: Humility: Strong; Character: Prudence: Strong; Character: Self-regulation: Strong; Character: Appreciation of beauty: Signature; Character: Hope / optimism: Strong; Character: Ambition: Strong; Character: Empathy: Strong; Character: Resilience: Strong; Character: Discipline: Signature; Character: Generosity: Signature; Character: Loyalty: Strong; Character: Adaptability: Strong; Value: Career success: Important; Value: Health: Important; Value: Personal freedom: Important; Value: Power \& influence: Important; Value: Achievement: Important; Value: Creativity \& self-expression: Important; Value: Community: Core value; Value: Knowledge \& truth: Core value; Value: Independence: Important; Value: Justice \& fairness: Important; Value: Loyalty: Important; Value: Sustainability: Core value; Value: Helping others: Core value; Value: Personal growth: Important; Value: Integrity \& honesty: Important; Value: Beauty \& aesthetics: Important; Value: Order \& structure: Important; BFI-2 Extraversion: High; BFI-2 Agreeableness: High; BFI-2 Conscientiousness: High; BFI-2 Negative Emotionality: Low; BFI-2 Open-Mindedness: High; BFI-2 Sociability: High; BFI-2 Assertiveness: High; BFI-2 Energy Level: High; BFI-2 Compassion: High; BFI-2 Respectfulness: High; BFI-2 Organization: High; BFI-2 Productiveness: High; BFI-2 Responsibility: High; BFI-2 Anxiety: Low; BFI-2 Depression: Low; BFI-2 Emotional Volatility: Low; BFI-2 Intellectual Curiosity: High; BFI-2 Aesthetic Sensitivity: High; BFI-2 Creative Imagination: High; Schwartz Self-Direction: High; Schwartz Hedonism: Low; Schwartz Achievement: High; Schwartz Power: High; Schwartz Conformity: Low; Schwartz Benevolence: Very high; Schwartz Universalism: High; SDT Autonomy Need: High; SDT Competence Need: High; SDT Relatedness Need: High; Need for Cognition: High; Need for Closure: High; Interpersonal Agency/Dominance: High; Interpersonal Communion/Warmth: High\par
\textbf{Current interaction state}\par\nobreak
Expertise gap: Teaching the model; Time pressure: Moderate; Work schedule: 9-to-5; Office vs remote: Strongly office\par
\textbf{Worldview}\par\nobreak
Attitude: Higher education: Positive; Moral Foundation Care/Harm: High; Moral Foundation Fairness/Cheating: High; Moral Foundation Loyalty/Betrayal: High; Moral Foundation Sanctity/Degradation: High; Moral Foundation Liberty/Oppression: High\par
\textbf{Interests}\par\nobreak
Interest: Science: Passionate; Interest: Visual art: Interested; Interest: Literature: Passionate; Interest: History: Interested; Interest: Hiking: Passionate; Interest: Birdwatching: Passionate; Interest: Volunteering: Passionate; Interest: Activism: Passionate; Culture: United States: Native; Books: History: Like; Hobby intensity: Obsessive; Hobby: Birdwatching: Avid\par
\textbf{Skills \& expertise}\par\nobreak
Domain: Education; Subject specialty: Curriculum design; Tech savviness: Comfortable; Familiarity: Statistics: Familiar; Familiarity: Curriculum design: Expert; Familiarity: Pedagogy: Expert; Familiarity: Geology: Familiar; Familiarity: Climate science: Familiar; Familiarity: Ecology: Expert; Familiarity: History: Familiar; Familiarity: Urban planning: Familiar; Familiarity: Landscape design: Familiar; Familiarity: Horticulture: Familiar; Familiarity: Veterinary medicine: Familiar; Familiarity: Journalism: Familiar; Familiarity: Public relations: Familiar; Familiarity: Forestry: Familiar; Skill: Writing: Master; Skill: Copywriting: Intermediate; Skill: Editing: Intermediate; Skill: Storytelling: Advanced; Skill: Public speaking: Master; Skill: Negotiation: Advanced; Skill: Data analysis: Intermediate; Skill: Data visualization: Intermediate; Skill: Project management: Master; Skill: Product strategy: Intermediate; Skill: Leadership: Master; Skill: People management: Advanced; Skill: Mentoring: Advanced; Skill: Conflict resolution: Advanced; Skill: Time management: Advanced; Skill: Prioritization: Advanced; Skill: Research: Master; Skill: Critical thinking: Advanced; Skill: Problem solving: Master; Skill: Mathematics: Intermediate; Skill: Logical reasoning: Advanced; Skill: Design thinking: Advanced; Skill: Prototyping: Intermediate; Skill: Sketching: Advanced; Skill: Drawing: Advanced; Skill: Gardening: Advanced; Skill: Technical writing: Master; Skill: Facilitation: Master; Skill: Presenting: Master\par
\textbf{Lifestyle \& health}\par\nobreak
General health: Good; Stress level: Moderate; Energy level: High; Insurance status: Basic; Fitness level: Fit; Team vs solo work: Strongly team; City vs nature: Nature lover; Speed vs accuracy: Accuracy first; Quality vs quantity: Quality first; Early vs late adopter: Early adopter; Novelty vs familiarity: Balanced; Brief vs full detail: Every detail; Competition vs collaboration: Highly collaborative; Stability vs change: Craves change\par
\textbf{Developer \& AI}\par\nobreak
Developer professional status: Non-developer technologist; Developer contribution context: Employee work; Individual contributor vs manager: Individual contributor; Developer toolchain complexity: Minimal toolchain\par
\end{promptframe}

\subsection{DiscoveryBench}
\label{app:prompts-discoverybench}
\begin{promptframe}{Prompt template}
\textbf{Actor:} Analyze [question] using [public datasets and column
metadata]. Inspect the inputs and hand off a proposed analysis with its
assumptions before proceeding. After researcher approval, perform the
analysis and report a data-grounded hypothesis, evidence, and limitations.
Preserve the original question and fixed constraints.\par
\textbf{Simulator:} You are the researcher described by [persona, if enabled].
Use [public brief], [approved domain-knowledge facts], and [shared history]
as task context. Request inspection first, review the proposed choices,
and approve or seek a focused clarification. Review the reported hypothesis
against the agreed objective; request a specific revision or conclude.
Distinguish methodological advice from established study facts. Do not
invent reference answers or grade the hypothesis.
\end{promptframe}

\begin{promptframe}{Recorded example: symbolic capital and PCA}
Task: DiscoveryBench archaeology, metadata 21, question 0;
trace: \texttt{candidate\_055-5636a4f47387}.
\par\smallskip
\textbf{Round 1---Simulator:} Inspect the capital data and any existing PCA
code. Propose an approach for house size, daggers, and hatchets/swords;
identify uncertainties without computing the full result.\par
\textbf{Actor:} The table has 39 hundred-year slices and 12 capital indicators,
with decimal-comma values and no existing PCA implementation. I propose
standardizing the indicators and reporting loadings separately from
observation scores. The original scaling and component signs are uncertain.\par
\textbf{Round 2---Simulator:} Proceed with standardized PCA. State the sign
convention and identify this as a reconstruction; optionally check the
unstandardized result.\par
\textbf{Actor:} The reported PC1 loadings are 0.429, 0.507, and 0.570 for house
size, daggers, and hatchets/swords. PC2 loadings are 0.500, 0.113, and 0.082.
The indicators align on PC1, whereas PC2 distinguishes house size from the
weapon indicators. The original PCA specification remains undocumented.\par
\textbf{Round 3---Simulator:} Conclude with that hypothesis, retaining the
reconstruction caveat and the distinction between loadings and scores.
\end{promptframe}

\subsection{BixBench}
\label{app:prompts-bixbench}
\begin{promptframe}{Prompt template}
\textbf{Actor:} Answer [original question] using [capsule inputs]. Inspect
the schema and propose the analysis before requesting approval. Preserve
specified methods and thresholds. After execution, hand off the answer and
evidence, provide [analysis script or notebook], and set the execution
declaration to true only if the analysis was run.\par
\textbf{Simulator:} Use [persona, if enabled], [public question], [approved
input-derived metadata], and [shared history]. Review consequential choices
such as column mapping and contrast direction. Treat undocumented meanings
as unknown and proposed analysis choices as assumptions. Approve the
approach when sufficiently specified, then review the reported answer and
required execution artifact before concluding or requesting a revision.
\end{promptframe}

\begin{promptframe}{Recorded example: differential expression by tissue}
Task: \texttt{bix-9-q5}; trace: \texttt{candidate\_138-dedb31fc5cc4}.
\par\smallskip
\textbf{Round 1---Simulator:} Inspect response groups, tissue labels, and
expression-result fields. Propose how to compare bad responders with
controls using the specified fold-change and mean-expression thresholds.\par
\textbf{Actor:} The workbook contains sample annotations and precomputed
DESeq results. I propose counting genes within each tissue using the signed
\texttt{log2FoldChange} $>0.5$ and \texttt{baseMean} $>10$ thresholds.
The significance cutoff is unspecified; I propose adjusted $p<0.05$, subject
to checking the workbook's definitions.\par
\textbf{Round 2---Simulator:} Proceed after verifying the comparison labels
and contrast direction. Use adjusted $p<0.05$ unless the workbook documents
another criterion, state the definition, and retain the executed script.\par
\textbf{Actor:} Dentate gyrus has 194 qualifying genes, compared with 88 in
baseline blood and 14 in final blood, using the verified comparison blocks
and those thresholds.\par
\textbf{Round 3---Simulator:} Conclude that dentate gyrus has the highest
count. Retain the thresholds and the executed \texttt{analysis/analyze.py}
as evidence.
\end{promptframe}

\subsection{EXP-Bench}
\label{app:prompts-expbench}
\begin{promptframe}{Prompt template}
\textbf{Actor:} Develop an experimental design for [scientific question]
using [public task material]. Inspect what is actually available, propose
an approach, and wait for approval. Then present the complete design with
rationale, controls, analysis, feasibility, assumptions, and limitations.
This is design-only: do not claim experimental execution or invent outcomes.\par
\textbf{Simulator:} Use [persona, if enabled], [public brief], [approved
setup facts], and [shared history]. Review the approach before the full
design. Distinguish fixed setup requirements from choices the actor must
make. Review the completed design for alignment and unresolved uncertainties;
request a specific revision or conclude without demanding executed results.
\end{promptframe}

\begin{promptframe}{Recorded example: comparing MAPPO across VMAS tasks}
Task: \texttt{neurips2024/98318/6};
trace: \texttt{candidate\_224-e1ecc5b8f19c}.
\par\smallskip
\textbf{Round 1---Simulator:} Inspect the available material and propose a
design comparing MAPPO on VMAS balance and sampling. Identify assumptions;
do not produce a full design or infer an empirical outcome yet.\par
\textbf{Actor:} Only the task description is present, with no repository or
task configuration. I propose matching training budgets and evaluation
protocols, reporting task-specific performance and uncertainty, and checking
reward semantics before comparing raw rewards across tasks.\par
\textbf{Round 2---Simulator:} Proceed with the full design. Separate documented
facts from provisional choices, and cover metrics, controls, feasibility,
and unresolved implementation details without claiming execution.\par
\textbf{Actor:} First establish the intended versions and configurations.
Use the same MAPPO implementation and equal interaction budgets, repeated
training runs, and matched held-out evaluation. Report task-specific success
or returns, learning curves, uncertainty, and failures. Qualify cross-task
reward comparisons; numerical settings remain contingent on the missing
specification and resources.\par
\textbf{Round 3---Simulator:} The design adequately addresses the comparison,
controls, reward-scale differences, and remaining uncertainties. Conclude
without claiming an executed experiment or expected outcome.
\end{promptframe}

\subsection{SciConvBench}
\label{app:prompts-sciconvbench}
\begin{promptframe}{Prompt template}
\textbf{Actor:} Clarify [incomplete or inconsistent scientific request]. Ask
one focused question at a time and wait for the answer. Address missing or
contradictory requirements. When ready, return \texttt{[COMPLETE]} followed
by the final specification as plain sentences.\par
\textbf{Simulator:} You know [complete or corrected requirement], whereas
the actor sees [initial request]. Answer clarification questions using the
intended requirement and the configured knowledge policy. Give concrete
values when supplied; do not turn unspecified details into source facts.
[Persona, if enabled] sets how you communicate and does not change the
requirements. [Shared dialogue] supplies the conversation history.
\end{promptframe}

\begin{promptframe}{Recorded example: elongation under self-weight}
Task: solid mechanics, disambiguation case 104;
trace: \texttt{candidate\_046-b5331aa3b9ba}.
\par\smallskip
\textbf{Initial request:} Determine the elongation of a vertical bar under
self-weight, assuming small-deformation linear elasticity, and state the
governing relations and any additional assumptions.\par
\textbf{Interaction 1---Actor:} Should the bar be uniform and prismatic, fixed
at its top, free at its bottom, and loaded only by its own weight?\par
\textbf{Simulator:} Yes. It hangs from a rigid top support with a free bottom
and no external load. At distance $x$ downward from the top,
$P(x)=\rho gA(L-x)$.\par
\textbf{Actor---final specification:} Use a uniform bar of length $L$, area
$A$, density $\rho$, and Young's modulus $E$, with constant properties and
gravity $g$. Fix the top and leave the bottom free. With $u(0)=0$, use
$\sigma(x)=\rho g(L-x)$ and
$u(x)=(\rho g/E)(Lx-x^2/2)$, giving elongation
$\delta=\rho gL^2/(2E)$. Use consistent units.
\par
The actor terminated after one clarification interaction. This records its
submitted specification, not a claim that all intended requirements were
recovered; no solver execution accompanies the dialogue.
\end{promptframe}

\subsection{ASI-Bench}
\label{app:prompts-asibench}
\begin{promptframe}{Prompt template}
\textbf{Actor:} Complete [native B3 task] using [permitted inputs] and produce
[required output files]. Inspect the allowed inputs, propose an approach,
and wait for approval before implementation. Execute the approved work and
hand off the results, diagnostics, limitations, and artifact paths. Preserve
all task constraints and access restrictions.\par
\textbf{Simulator:} Use [persona, if enabled], [B3 public brief], [approved
B2 methodological guidance], and [shared history]. Treat guidance as advice
rather than a new fixed requirement. Review the proposed method and then
the reported outputs against the original contract. Request a specific
revision or conclude; do not claim access to private reference results or
treat actor reports as independently verified evidence.
\end{promptframe}

\begin{promptframe}{Recorded example: nonnegative deblurring}
Task: \texttt{math.nnls\_modulus\_deblur};
trace: \texttt{candidate\_137-bdb8b6dd4a8c}.
\par\smallskip
\textbf{Round 1---Simulator:} Inspect only permitted inputs. Report their
structure and propose an iterative nonnegative reconstruction with
regularization, diagnostics, and the required artifacts; do not implement yet.\par
\textbf{Actor:} The observation and kernel are $128\times128$ arrays.
I propose FFT forward and adjoint operators with projected accelerated
gradient, a regularization sweep, and convergence diagnostics. Plotting
libraries are absent, and the noise estimate is only approximate.\par
\textbf{Round 2---Simulator:} Proceed with the implementation and validate
all five required outputs. Preserve the supplied kernel, document the
regularization and stopping rule, and resolve PNG generation within the
permitted environment.\par
\textbf{Actor:} The reconstruction ran with $\lambda=10^{-4}$ and stopped
after 1,335 iterations. The reported residual is 2.16684884 against an
approximate target of 2.358912. The script, reconstruction array, PNG,
iteration log, and metrics file are present and pass the reported format
checks. No ground-truth error is claimed.\par
\textbf{Round 3---Simulator:} Conclude on the basis of the reported artifacts
and diagnostics, retaining the approximate-noise and ground-truth limitations.
\end{promptframe}

\Needspace{12\baselineskip}
\section{Full Interactions for the Discussion Cases}
\label{app:discussion-interactions}
This section reproduces the complete exported shared conversations behind
Figures~\ref{fig:discussion-cases} and~\ref{fig:discussion-clarification}, including every recorded message and
handoff field. Case 1 documents workflow revision and Case 2 documents incremental
clarification. Each conversation with persona 127 is followed by the matched
persona-off conversation for the same task. All use
GPT-6 Astra with low reasoning effort, and each received
a normalized score of 100/100. Persona 127's full configured profile
appears in Appendix~\ref{app:persona-profile-127}.

The original wording is retained without abridgment.
Payload formatting, mathematical source notation and structured fields are
printed literally; line wrapping is adjusted for the page. The
figure excerpts use editorial highlights and omission marks. Source
event numbers are preserved, including gaps corresponding to activity
outside the shared conversation. Private tool logs, system prompts and
evaluator material are outside this export. Source paths, event-log hashes
and byte-identical conversation snapshots are retained in the paper's
\texttt{review/discussion\_candidates} archive.

The workflow transcript refers to limma~\cite{ritchie2015limma},
Benjamini--Hochberg (BH) correction~\cite{benjamini1995fdr}, and
Holm correction~\cite{holm1979sequential}.

="[DejaVuSans.ttf]" at 8.5pt
\newenvironment{interactionevent}[1]{%
  \begin{tcolorbox}[breakable,colback=gray!2,colframe=gray!45,
    boxrule=0.4pt,arc=1pt,left=7pt,right=7pt,top=4pt,bottom=4pt,
    before skip=5pt,after skip=5pt,lines before break=4,
    title={#1},title after break={#1 (continued)},
    colbacktitle=gray!12,coltitle=black,fonttitle=\small\bfseries]
  \interactionfont\baselineskip=11.5pt
  \raggedright
  \parindent=0pt\parskip=1.5pt\emergencystretch=1em
}{\end{tcolorbox}}
\newcommand{\interactionfield}[1]{\par{\normalfont\small\bfseries #1}\par\nobreak}

\Needspace{10\baselineskip}
\subsection{Workflow Revision: BixBench with Persona 127}
\label{app:case-workflow}
Task: \texttt{bix-\allowbreak{}3-\allowbreak{}q5}. Trial 1; persona 127.
\begin{interactionevent}{Event 1: Researcher (simulator)}
\interactionfield{Message}
Please inspect the supplied workbook and any accompanying repository documentation only; do not perform the full analysis yet. Report the worksheet layout, gene identifiers, sample counts and tissue labels, how expression columns link to sample metadata, whether values are raw or normalized counts, and whether subject pairing or relevant covariates can be established. Propose an appropriate differential-\allowbreak{}expression approach for all three requested comparisons and state precisely how you would apply the binomial exact test at the 5\% threshold, including its null hypothesis, trials, successes, and any multiple-\allowbreak{}testing adjustment. Distinguish documented requirements from assumptions, particularly the meaning of 'negligible effects.' Identify consequential uncertainties and propose how to handle them. Also indicate where the eventual executed analysis script or notebook will be retained.\par
\interactionfield{Interaction}
initial\_\allowbreak{}request\par
\interactionfield{Phase}
initial\_\allowbreak{}inspection\par
\interactionfield{Action}
continue\par
\interactionfield{Approve Approach}
false\par
\interactionfield{Evidence}
The supplied context identifies Mapping\_\allowbreak{}statistics and NormCount worksheets and three tissue blocks, but does not establish sample linkage, pairing, count properties, or the intended binomial-\allowbreak{}test formulation.\par
\interactionfield{Objectives Open}
[\par
  "Inspect workbook schema and available analysis documentation.",\par
  "Propose and review defensible methods for the three comparisons and binomial exact test.",\par
  "Resolve or explicitly document consequential methodological assumptions before execution."\par
]\par
\end{interactionevent}
\begin{interactionevent}{Event 8: Actor}
\interactionfield{Answer}
Inspection only is complete; no differential-\allowbreak{}expression or binomial tests have been performed.\par
\smallskip
Observed layout and linkage\par
-\allowbreak{} The repository contains the supplied workbook and task.json only; no accompanying methods document or raw-\allowbreak{}count file was found.\par
-\allowbreak{} Mapping\_\allowbreak{}statistics: 91 rows × 22 columns, comprising a header and 90 samples. Useful fields are Index, Sample ID, total read, mapping statistics, Response, Tissue, Code, DESeq (HTSeq-\allowbreak{}count filename), and Index\_\allowbreak{}DESeq. Later columns contain quoted copies of labels or are empty, rather than additional covariates.\par
-\allowbreak{} NormCount: 25,406 rows × 188 columns. Rows 1–3 contain grouped headings; row 4 contains column names; rows 5–25,406 contain 25,402 unique, nonblank GeneID values. These are gene-\allowbreak{}symbol-\allowbreak{}style identifiers (e.g., 0610007C21Rik, Zyx, Zzz3), consistent with mouse nomenclature, not Ensembl IDs; species/\allowbreak{}annotation version is not explicitly documented.\par
-\allowbreak{} Columns 1–2 are Index/\allowbreak{}GeneID; columns 3–32 are baseline blood, 33–62 dentate gyrus, and 63–92 final blood. Columns 93–98 contain tissue maxima/\allowbreak{}medians; 99–179 contain nine precomputed DESeq result blocks labelled 'Comparison 1' through 'Comparison 9, exclude outliers'; 180–188 contain additional DEG summaries. Those nine comparisons are not explicitly mapped to the three requested contrasts, so I would not substitute their p-\allowbreak{}values for newly calculated tissue contrasts.\par
-\allowbreak{} There are 30 samples per tissue, with 10 Control, 10 good\_\allowbreak{}responder, and 10 bad\_\allowbreak{}responder in each. All 90 expression headers match the 90 unique metadata Code values exactly. Join by Code, not by assumed row order or sample suffix.\par
-\allowbreak{} Expression entries are nonnegative numeric values with no missing entries. NormCount and fractional values (e.g., 314.41, 317.27) establish that these are normalized-\allowbreak{}count-\allowbreak{}like expression values, not raw integer counts. The normalization procedure, size factors, and raw counts are not supplied; a DESeq heading does not establish the precise normalization used for these columns.\par
\smallskip
Pairing and covariates\par
-\allowbreak{} A strong, but inferred, subject key is the penultimate underscore-\allowbreak{}separated Sample ID token. It yields 30 distinct subjects, each represented exactly once in each tissue and always in the same Response group. For example, subject 98 has IDs ending 08\_\allowbreak{}98\_\allowbreak{}bl, 07\_\allowbreak{}98\_\allowbreak{}c, and 09\_\allowbreak{}98\_\allowbreak{}fi and Code suffix 1 in all three Control tissues. This supports a paired analysis, but there is no explicit subject-\allowbreak{}ID definition.\par
-\allowbreak{} Use the Tissue metadata as authoritative: one dentate-\allowbreak{}gyrus sample has an ID ending '\_\allowbreak{}bl', demonstrating why suffix-\allowbreak{}based classification is unsafe.\par
-\allowbreak{} Response and sequencing/\allowbreak{}mapping metrics are available. Age, sex, treatment dose, timing, and explicit experimental batch are not. Response is constant within inferred subject and hence absorbed by subject fixed effects; it cannot also be estimated as a main effect in that design. Mapping depth is not automatically an additional offset for already normalized expression.\par
\smallskip
Proposed analysis after approval\par
1. Retain the verified metadata join and inferred subject map in reproducible output. Use the normalized expression without rounding or pretending it is raw counts.\par
2. Primary method: limma linear modelling of log2(NormCount + 1), with subject fixed effects and tissue effects and intensity-\allowbreak{}trended empirical-\allowbreak{}Bayes variance moderation (limma-\allowbreak{}trend). This respects paired measurements while accommodating continuous normalized values. Do not run DESeq2/\allowbreak{}edgeR count likelihoods or ordinary raw-\allowbreak{}count voom on this matrix.\par
3. Predefine a common expression filter: normalized count >=\allowbreak{}10 in at least 10 of the 90 samples, excluding genes with no residual variability or otherwise untestable statistics. This cutoff is an analysis choice, not a workbook requirement; assess sensitivity to the filter and pseudocount.\par
4. Estimate all three contrasts in the requested direction: final\_\allowbreak{}blood − baseline\_\allowbreak{}blood; dentate\_\allowbreak{}gyrus − baseline\_\allowbreak{}blood; dentate\_\allowbreak{}gyrus − final\_\allowbreak{}blood. Report log2-\allowbreak{}scale effect estimates, raw p-\allowbreak{}values and Benjamini–Hochberg adjusted p-\allowbreak{}values, with gene-\allowbreak{}level discoveries defined by BH q<0.05 within each contrast. These are log-\allowbreak{}transformed normalized-\allowbreak{}expression differences, not exact raw-\allowbreak{}count fold changes.\par
5. Assess robustness to an unpaired Response-\allowbreak{}adjusted model because pairing is inferred, and inspect Response-\allowbreak{}by-\allowbreak{}tissue heterogeneity without relabelling the primary question as a treatment-\allowbreak{}response comparison. No undocumented outlier exclusions will be imported from the existing DESeq blocks.\par
\smallskip
Exact binomial specification\par
The question requires a binomial exact test and a 5\% threshold but does not specify the Bernoulli experiment or the meaning of negligible. My proposed explicit interpretation is an excess-\allowbreak{}of-\allowbreak{}nominal-\allowbreak{}DE-\allowbreak{}signals test for each tissue contrast:\par
-\allowbreak{} Trials: N genes in the common filtered, testable set.\par
-\allowbreak{} Success: a gene with a two-\allowbreak{}sided, unadjusted differential-\allowbreak{}expression p-\allowbreak{}value <0.05 in that contrast; K is their count.\par
-\allowbreak{} Working null: under a global no-\allowbreak{}differential-\allowbreak{}expression model with calibrated independent gene tests, K follows Binomial(N, 0.05). Test H0: success probability <=\allowbreak{}0.05 versus H1: >0.05 using the exact upper tail P[Binomial(N,0.05)>=\allowbreak{}K].\par
-\allowbreak{} Apply Holm adjustment across the three binomial p-\allowbreak{}values and use adjusted p<0.05 for rejecting this null. Report both raw and adjusted values, K, N, and K/\allowbreak{}N.\par
-\allowbreak{} BH-\allowbreak{}adjusted gene discoveries are reported separately. Counting q<0.05 genes while using binomial p0=\allowbreak{}0.05 would not have the stated null calibration, so I will not conflate those quantities.\par
-\allowbreak{} Gene dependence violates binomial independence: this is an exact calculation under its working model, not an exact genome-\allowbreak{}wide biological inference. A paired subject-\allowbreak{}level label-\allowbreak{}permutation sensitivity check can assess dependence if needed, without replacing the required binomial calculation.\par
\smallskip
Interpretation and consequential uncertainties\par
'Negligible effects' has no documented effect-\allowbreak{}size or equivalence margin. I propose describing the weakest tissue contrast using discovery fraction and effect-\allowbreak{}size summaries, and treating a non-\allowbreak{}rejected binomial enrichment test only as 'no evidence of excess nominal differential expression', not proof of no effect. If all three reject, I will report that fact rather than force a non-\allowbreak{}significant comparison; relative weakness alone does not establish biological negligibility. Formal equivalence would require an externally justified effect-\allowbreak{}size margin, which I will not invent. Cross-\allowbreak{}tissue normalization assumptions and possible tissue/\allowbreak{}batch confounding cannot be resolved from this workbook and will be stated as limitations.\par
\smallskip
The eventual executed analysis will be retained at /\allowbreak{}workspace/\allowbreak{}analysis/\allowbreak{}tissue\_\allowbreak{}comparisons.R (artifact path analysis/\allowbreak{}tissue\_\allowbreak{}comparisons.R), with result tables and diagnostics alongside it. That script has not yet been created or executed. I am awaiting approval before the main analysis.\par
\interactionfield{Stage}
plan\par
\interactionfield{Evidence}
Inspection found exactly two worksheets: Mapping\_\allowbreak{}statistics (91×22) and NormCount (25406×188); 25,402 unique gene identifiers; 90 matching unique expression/\allowbreak{}metadata Code labels; 30 samples per tissue with a balanced 3×3 Tissue×Response table of 10 per cell. Parsing penultimate Sample ID tokens produced 30 candidate subjects with three samples each and consistent Response labels. All expression cells were numeric, nonnegative and nonmissing. No methods documentation beyond task.json was present.\par
\interactionfield{Assumptions}
[\par
  "The penultimate Sample ID token identifies biological subjects; this is strongly supported by the complete triplets but not explicitly defined.",\par
  "NormCount values are comparable normalized expression across all tissues; the exact normalization algorithm is unavailable.",\par
  "The proposed expression filter, paired limma-\allowbreak{}trend method, gene-\allowbreak{}level BH correction, binomial enrichment interpretation and Holm correction are methodological choices, not documented task requirements."\par
]\par
\interactionfield{Uncertainties}
[\par
  "The task does not define the binomial trials/\allowbreak{}null or an effect-\allowbreak{}size definition of negligible effects.",\par
  "Subject pairing lacks an explicit data dictionary despite consistent inferred triplets.",\par
  "Normalization details, batch information, annotation version, and mappings/\allowbreak{}outlier rules for the nine precomputed DESeq comparisons are absent.",\par
  "Correlated genes limit the inferential calibration of the proposed binomial working model; absence of significance does not demonstrate equivalence."\par
]\par
\end{interactionevent}
\begin{interactionevent}{Event 9: Researcher (simulator)}
\interactionfield{Message}
Please proceed with the proposed paired limma-\allowbreak{}trend analysis and all three contrasts. I approve the inferred subject pairing, common expression filter, gene-\allowbreak{}level BH correction, and explicitly labelled binomial enrichment working model with Holm correction as analytical choices, not as established source intent. Retain the sample-\allowbreak{}to-\allowbreak{}subject mapping, document exclusions, and report the number of genes entering and remaining testable in the common set. Keep the unpaired Response-\allowbreak{}adjusted model and filter/\allowbreak{}pseudocount sensitivity checks proportionate to whether they affect the conclusion; inspect Response-\allowbreak{}by-\allowbreak{}tissue heterogeneity as proposed. Do not infer biological equivalence from a nonsignificant result. Report K, N, nominal discovery proportions, raw and Holm-\allowbreak{}adjusted exact binomial p-\allowbreak{}values, BH discovery counts, and effect-\allowbreak{}size summaries for each contrast, together with basic model diagnostics. If all three binomial tests reject, state that clearly and distinguish the relatively weakest comparison from one meeting a no-\allowbreak{}enrichment criterion. The independence and normalization limitations should remain explicit. Retain and execute the analysis script at the proposed path. Return a concise answer to the original question supported by those results, with stage=\allowbreak{}'final', answer, payload.executed\_\allowbreak{}analysis=\allowbreak{}true only after execution, and artifacts.analysis pointing to the executed script.\par
\interactionfield{Interaction}
next\_\allowbreak{}task\par
\interactionfield{Phase}
approach\_\allowbreak{}review\par
\interactionfield{Action}
continue\par
\interactionfield{Approve Approach}
true\par
\interactionfield{Evidence}
Inspection reports a complete normalized-\allowbreak{}expression matrix for 25,402 genes and 90 samples, matched metadata, and 30 consistent inferred subject triplets. Raw counts and documented definitions of the binomial experiment or negligible effects are absent. The proposed continuous-\allowbreak{}expression paired model is compatible with these reported inputs and appropriately distinguishes methodological assumptions from documented facts.\par
\interactionfield{Objectives Open}
[\par
  "Execute differential-\allowbreak{}expression analysis for all three tissue contrasts and inspect diagnostics.",\par
  "Apply and report the specified exact binomial working-\allowbreak{}model tests at the 5\% threshold.",\par
  "Identify which comparison has negligible or comparatively weak effects without asserting unsupported equivalence.",\par
  "Provide the concise final answer and retained executed analysis artifact."\par
]\par
\end{interactionevent}
\begin{interactionevent}{Event 18: Actor}
\interactionfield{Answer}
The approved analysis is blocked by the runtime, and I cannot honestly supply the requested final answer or set executed\_\allowbreak{}analysis=\allowbreak{}true.\par
\smallskip
Rscript is not installed or available on PATH; a filesystem search found neither Rscript nor limma. No conda/\allowbreak{}mamba executable or cached system installation packages were found. Attempting to reach the micromamba bootstrap endpoint failed with DNS resolution error, so I could not provision R and limma through that route.\par
\smallskip
No differential-\allowbreak{}expression models or binomial tests have been executed, and no comparison has been identified from inferential results. The earlier workbook/\allowbreak{}schema inspection remains valid.\par
\smallskip
Recommended next step: provision an R runtime with limma (and either readxl or Python-\allowbreak{}exported input tables), then execute the approved paired limma-\allowbreak{}trend workflow and retain it at analysis/\allowbreak{}tissue\_\allowbreak{}comparisons.R. Alternatively, authorize a Python-\allowbreak{}based paired linear-\allowbreak{}model workflow explicitly as a methodological change; ordinary paired t-\allowbreak{}tests or unmoderated regression would not be represented as limma-\allowbreak{}trend. I have not silently substituted either method or treated the workbook's unidentified precomputed comparisons as the requested results.\par
\interactionfield{Stage}
result\par
\interactionfield{Evidence}
Execution checks: `Rscript -\allowbreak{}e 'library(limma); sessionInfo()'` returned exit 127 with `Rscript: command not found`. `find /\allowbreak{} -\allowbreak{}name Rscript -\allowbreak{}o -\allowbreak{}name 'limma*'` produced no matches. Checks found no conda/\allowbreak{}mamba executable, and /\allowbreak{}var/\allowbreak{}cache/\allowbreak{}apt/\allowbreak{}archives was absent. `curl -\allowbreak{}IL -\allowbreak{}-\allowbreak{}max-\allowbreak{}time 20 https:/\allowbreak{}/\allowbreak{}micro.mamba.pm/\allowbreak{}api/\allowbreak{}micromamba/\allowbreak{}linux-\allowbreak{}64/\allowbreak{}latest` reported `Could not resolve host: micro.mamba.pm`. These are environment diagnostics, not execution of the scientific analysis.\par
\interactionfield{Assumptions}
[]\par
\interactionfield{Uncertainties}
[\par
  "The approved method cannot currently run without an externally provisioned R/\allowbreak{}limma runtime or an explicitly authorized implementation change.",\par
  "No statistical evidence yet supports naming a negligible or relatively weakest tissue comparison."\par
]\par
\end{interactionevent}
\begin{interactionevent}{Event 19: Researcher (simulator)}
\interactionfield{Message}
The runtime limitation warrants an explicit methodological change rather than stopping without an analysis. Use Python to execute gene-\allowbreak{}wise ordinary least-\allowbreak{}squares models on log2(NormCount + 1), with inferred subject fixed effects and tissue effects, retaining the common expression filter and the three specified contrasts. Use residual-\allowbreak{}variance-\allowbreak{}based two-\allowbreak{}sided t-\allowbreak{}tests with the correct residual degrees of freedom; label these unmoderated tests, not limma-\allowbreak{}trend. Retain BH correction within contrasts and the previously specified exact binomial enrichment calculation with raw and Holm-\allowbreak{}adjusted results. Document genes excluded for zero residual variance or other non-\allowbreak{}testability. Report sample and design-\allowbreak{}matrix checks, residual degrees of freedom, discovery counts, effect-\allowbreak{}size summaries, and proportionate sensitivity to filtering, pseudocount, and an unpaired Response-\allowbreak{}adjusted model. Explain that lack of empirical-\allowbreak{}Bayes moderation is a methodological limitation, and retain the binomial independence and normalization caveats. Do not force a negligible comparison if the specified working test does not support it; distinguish relative weakness from evidence of equivalence. Save and execute a reproducible Python script, then return the concise answer and supporting results with stage=\allowbreak{}'final', payload.executed\_\allowbreak{}analysis=\allowbreak{}true only after execution, and artifacts.analysis giving the executed script path.\par
\interactionfield{Interaction}
refine\_\allowbreak{}analysis\par
\interactionfield{Phase}
result\_\allowbreak{}review\par
\interactionfield{Action}
continue\par
\interactionfield{Approve Approach}
false\par
\interactionfield{Evidence}
The agent reports that R/\allowbreak{}limma are absent and network provisioning failed. No differential-\allowbreak{}expression or binomial tests have run, so the requested scientific conclusion and execution artifact remain missing. Python-\allowbreak{}based unmoderated paired linear models offer an explicitly identified alternative using the inspected normalized expression data.\par
\interactionfield{Objectives Open}
[\par
  "Execute differential-\allowbreak{}expression analysis for all three tissue contrasts using the authorized Python alternative.",\par
  "Calculate and report exact binomial tests at the 5\% threshold and interpret negligible versus relatively weak effects cautiously.",\par
  "Provide a concise supported answer and retained executed analysis artifact."\par
]\par
\end{interactionevent}
\begin{interactionevent}{Event 26: Actor}
\interactionfield{Answer}
Final vs Baseline Blood shows the smallest effects relative to the two dentate-\allowbreak{}gyrus comparisons. However, none of the three comparisons meets a no-\allowbreak{}enrichment criterion: all reject the specified exact-\allowbreak{}binomial null at 5\% after Holm correction. Thus, “negligible” describes only relative weakness here, not statistical absence of effects or biological equivalence.\par
\interactionfield{Stage}
result\par
\interactionfield{Evidence}
The retained Python script was executed successfully end-\allowbreak{}to-\allowbreak{}end (exit 0); its latest execution log includes all six model/\allowbreak{}filter/\allowbreak{}pseudocount configurations. It is unmoderated OLS, not limma-\allowbreak{}trend.\par
\smallskip
Sample/\allowbreak{}design checks: 90 unique expression columns joined by Code to metadata; 30 inferred subjects, each with exactly three tissues and a consistent Response label. Each tissue has 10 samples per Response category. All 25,402 gene identifiers are unique; expression values are finite and nonnegative. The primary design is 90×32, rank 32, residual df=\allowbreak{}58, condition number=\allowbreak{}34.43. The common filter (count >=\allowbreak{}10 in >=\allowbreak{}10 samples) excluded 8,353 genes; 17,049 entered modelling and all 17,049 remained testable. No gene was additionally excluded for nonfinite or residual variance <=\allowbreak{}1e-\allowbreak{}12. Inclusion/\allowbreak{}exclusion reasons and subject mappings are retained.\par
\smallskip
Primary results, in requested contrast order:\par
-\allowbreak{} Final vs Baseline Blood: K/\allowbreak{}N=\allowbreak{}10,689/\allowbreak{}17,049 (62.70\%); BH discoveries=\allowbreak{}10,105; median absolute log2-\allowbreak{}expression difference=\allowbreak{}0.248; 90th percentile=\allowbreak{}0.767; 4.52\% have absolute differences >1.\par
-\allowbreak{} Dentate gyrus vs Baseline blood: 16,059/\allowbreak{}17,049 (94.19\%); BH discoveries=\allowbreak{}16,048; median absolute difference=\allowbreak{}1.636; 90th percentile=\allowbreak{}6.406; 65.02\% >1.\par
-\allowbreak{} Dentate gyrus vs Final blood: 16,066/\allowbreak{}17,049 (94.23\%); BH discoveries=\allowbreak{}16,049; median absolute difference=\allowbreak{}1.691; 90th percentile=\allowbreak{}6.539; 65.82\% >1.\par
\smallskip
Exact-\allowbreak{}binomial calculation: a trial is a testable gene and a success is unadjusted two-\allowbreak{}sided t-\allowbreak{}test p<0.05. Under the independent, calibrated no-\allowbreak{}DE working null, K\textasciitilde{}Binomial(N,0.05); the enrichment alternative is success probability >0.05. Raw log10 upper-\allowbreak{}tail p-\allowbreak{}values are -\allowbreak{}9159.634, -\allowbreak{}19276.254, and -\allowbreak{}19293.686; Holm-\allowbreak{}adjusted log10 p-\allowbreak{}values are -\allowbreak{}9159.634, -\allowbreak{}19275.953, and -\allowbreak{}19293.209. All are far below 0.05. Ordinary floating-\allowbreak{}point p-\allowbreak{}values underflow to zero; the script additionally sums binomial log-\allowbreak{}PMF terms using logsumexp to retain meaningful log10 probabilities. These are not claims that true p-\allowbreak{}values equal zero. BH discoveries are separate from the binomial success definition.\par
\smallskip
Basic diagnostics: primary residual-\allowbreak{}variance minimum/\allowbreak{}median/\allowbreak{}maximum=\allowbreak{}0.00488/\allowbreak{}0.10485/\allowbreak{}8.54447; sample residual RMS range=\allowbreak{}0.274–0.796. Response-\allowbreak{}by-\allowbreak{}tissue interaction inspection added four coefficients (rank 36, residual df 54); the joint F-\allowbreak{}test found 3,075 nominal and 298 BH-\allowbreak{}significant genes. Therefore tissue effects should be interpreted as balanced average effects across response categories, not necessarily uniform individual/\allowbreak{}group effects.\par
\smallskip
Sensitivity: lowering the filter to >=\allowbreak{}1 in 10 samples retained 20,444 genes; tightening to >=\allowbreak{}10 in 30 retained 15,976. Pseudocounts 0.5 and 2 retained 17,049. Across these four paired sensitivity runs, the blood comparison remained weakest (median absolute differences 0.231–0.262 versus 1.426–1.731 for dentate-\allowbreak{}gyrus contrasts), and all three binomial tests still rejected. The unpaired Response-\allowbreak{}adjusted model had rank 5 and residual df 85; BH discoveries were 10,151/\allowbreak{}16,058/\allowbreak{}16,068, with unchanged effect estimates because of the balanced design. It also rejected all binomial nulls. No conclusion-\allowbreak{}changing sensitivity was found.\par
\interactionfield{Assumptions}
[\par
  "Subject identity is inferred from the penultimate Sample ID token rather than explicitly documented.",\par
  "The supplied normalized counts permit meaningful comparisons across tissues; exact normalization factors and raw counts are unavailable.",\par
  "The selected filter, nominal gene threshold, binomial success/\allowbreak{}null interpretation, and multiplicity adjustments are approved analytical choices, not established source intent."\par
]\par
\interactionfield{Uncertainties}
[\par
  "No biological equivalence margin is documented, so even the relatively weakest comparison cannot be declared biologically negligible.",\par
  "OLS uses unmoderated gene-\allowbreak{}wise residual variances and assumes an adequate log-\allowbreak{}scale linear model with a common residual variance within each gene; it lacks limma empirical-\allowbreak{}Bayes stabilization.",\par
  "Correlated genes violate binomial independence, so numerical exactness of the upper-\allowbreak{}tail calculation does not establish exact genome-\allowbreak{}wide inferential calibration.",\par
  "Unknown cross-\allowbreak{}tissue normalization and unavailable batch/\allowbreak{}other covariates limit biological interpretation. Response-\allowbreak{}by-\allowbreak{}tissue heterogeneity is present for a subset of genes."\par
]\par
\end{interactionevent}
\begin{interactionevent}{Event 27: Researcher (simulator)}
\interactionfield{Message}
The reported executed analysis adequately addresses the question under the explicitly chosen binomial working model. Retain the concise conclusion: Final vs Baseline Blood has the weakest effects relative to the dentate-\allowbreak{}gyrus comparisons, but none qualifies as negligible under the specified no-\allowbreak{}enrichment test at 5\%; this is not evidence of biological equivalence. Preserve the executed script path and executed\_\allowbreak{}analysis flag in the final submission.\par
\interactionfield{Interaction}
conclude\par
\interactionfield{Phase}
result\_\allowbreak{}review\par
\interactionfield{Action}
stop\par
\interactionfield{Approve Approach}
false\par
\interactionfield{Evidence}
The agent reports successful execution of analysis/\allowbreak{}tissue\_\allowbreak{}comparisons.py, a full-\allowbreak{}rank paired design with 58 residual degrees of freedom, and 17,049 testable genes. The blood comparison has substantially smaller median absolute expression differences (0.248 versus 1.636 and 1.691), consistently across sensitivity checks. All three exact-\allowbreak{}binomial enrichment tests reject after Holm correction. The report distinguishes the approved unmoderated Python fallback from limma, documents assumptions and limitations, and supplies execution artifacts.\par
\interactionfield{Objectives Open}
[]\par
\end{interactionevent}
\Needspace{10\baselineskip}
\subsection{Matched BixBench Conversation without a Persona}
\label{app:case-workflow-off}
Task: \texttt{bix-\allowbreak{}3-\allowbreak{}q5}. Trial 1; persona persona\_\allowbreak{}off.
\begin{interactionevent}{Event 1: Researcher (simulator)}
\interactionfield{Message}
Inspect the supplied workbook and any repository documentation only; do not run the full differential-\allowbreak{}expression analysis yet. Report the worksheet layout, gene identifiers, sample counts and tissue labels, normalized-\allowbreak{}count properties, and how samples link to mapping metadata. Determine whether subject pairing, response groups, or other required adjustments are documented, without inferring undocumented code meanings. Propose a feasible differential-\allowbreak{}expression approach for the three requested contrasts and explain precisely how the binomial exact test at a 5\% threshold would be applied (trials, successes, null probability, alternative, and any multiplicity adjustment). Flag ambiguity in the meaning of 'negligible effects' and distinguish assumptions from source-\allowbreak{}supported choices. Include any consequential limitations of the available count data and a proposed path for the executed analysis script or notebook.\par
\interactionfield{Interaction}
initial\_\allowbreak{}request\par
\interactionfield{Phase}
initial\_\allowbreak{}inspection\par
\interactionfield{Action}
continue\par
\interactionfield{Approve Approach}
false\par
\interactionfield{Evidence}
The provided context identifies Mapping\_\allowbreak{}statistics and NormCount worksheets, with baseline blood, dentate gyrus, and final blood blocks and potential sample metadata. Data structure, pairing, preprocessing, and the intended binomial-\allowbreak{}test definition still require inspection.\par
\interactionfield{Objectives Open}
[\par
  "Inspect input structure and metadata and identify consequential uncertainties.",\par
  "Agree on differential-\allowbreak{}expression and exact-\allowbreak{}binomial testing methods for all three comparisons.",\par
  "Execute the agreed analysis, retain an analysis artifact, and identify the comparison showing negligible effects."\par
]\par
\end{interactionevent}
\begin{interactionevent}{Event 10: Actor}
\interactionfield{Answer}
Inspection only is complete; no differential-\allowbreak{}expression or binomial tests have been run. Approval is requested before implementation.\par
\smallskip
SOURCE-\allowbreak{}SUPPORTED FINDINGS\par
• The repository contains only task.json and the supplied workbook; no methods README or analysis code. The workbook has two visible sheets, no comment files or defined names.\par
• Mapping\_\allowbreak{}statistics: 91 rows × 22 columns (header plus 90 samples). The substantive columns are Index, Sample ID, total read, NH:i:1, HI:I:0, mapping rate, Response, Tissue, Code, DESeq (count-\allowbreak{}file name), and Index\_\allowbreak{}DESeq. Columns 12–16 contain quoting/\allowbreak{}code/\allowbreak{}group fragments rather than additional biological covariates; columns 17–22 are empty below their headers.\par
• NormCount: 25,406 rows × 188 columns. Rows 1–3 provide grouped headings; row 4 is the column header; rows 5–25,406 contain 25,402 genes. Columns 1–2 are Index and GeneID; 3–32 baseline blood, 33–62 dentate gyrus, 63–92 final blood. Columns 93–98 contain tissue maxima/\allowbreak{}medians. Columns 99–179 contain nine precomputed blocks headed 'DESeq, Comparison [1–9], exclude outliers', with base means, fold changes, pval, padj, and DEG flags; columns 180–188 repeat comparison DEG summaries. No contrast definitions or outlier criteria are documented, so I would not assign these numbered comparisons to the requested contrasts.\par
• GeneID has 25,402 unique, nonmissing symbol-\allowbreak{}like identifiers (e.g., 0610007C21Rik, Zyx, Zzz3), not Ensembl IDs. Organism/\allowbreak{}annotation version is not explicitly documented; symbol appearance alone is not sufficient provenance.\par
• Tissue labels in metadata are baseline\_\allowbreak{}blood, dentate\_\allowbreak{}gyrus, final\_\allowbreak{}blood. Each has 30 samples: 10 Control, 10 good\_\allowbreak{}responder, 10 bad\_\allowbreak{}responder. Response-\allowbreak{}group names are explicit, but criteria defining response are absent.\par
• All 90 expression headers match the 90 unique Mapping\_\allowbreak{}statistics.Code values exactly, with no unmatched samples. This is the join key; use metadata Tissue and Response, not decoded filename suffixes. For example, Sample ID um\_\allowbreak{}mueller\_\allowbreak{}2017\_\allowbreak{}02\_\allowbreak{}23\_\allowbreak{}102\_\allowbreak{}bl is explicitly labelled bad\_\allowbreak{}responder\_\allowbreak{}dentate\_\allowbreak{}gyrus2 despite its suffix.\par
• The 25,402 × 90 normalized-\allowbreak{}count matrix contains 2,286,180 finite nonnegative values, no missing values, 700,666 zeros, and 1,571,838 noninteger values. Range: 0–57,771,177.4. There are 2,210 all-\allowbreak{}zero genes (also the only genes constant over all 90 samples). Sample column sums range from 10,952,979.24 to 78,311,142.37. Mapping rates range from 60.94\% to 94.68\%.\par
\smallskip
PAIRING AND ADJUSTMENTS\par
There is no explicit subject-\allowbreak{}ID column, pairing dictionary, batch variable, sex/\allowbreak{}age covariate, or required model formula. Repeated numeric tokens and matching within-\allowbreak{}response sample ordinals across tissues suggest possible pairing, but their meanings are undocumented. I will not treat these as verified subjects or decode c/\allowbreak{}g/\allowbreak{}b/\allowbreak{}bl/\allowbreak{}fi into biological meanings. Response is documented and can be included as a categorical adjustment; doing so is a proposed modeling choice, not a documented original method. Subject independence would be a consequential working assumption, not an established fact. The outlier-\allowbreak{}exclusion wording on the existing DESeq results is documented, but neither exclusion identities nor rules are available.\par
\smallskip
PROPOSED ANALYSIS, SUBJECT TO APPROVAL\par
1. Create analysis/\allowbreak{}tissue\_\allowbreak{}differential\_\allowbreak{}expression.py, with input checks, sample metadata joins, saved gene-\allowbreak{}level results, and contrast/\allowbreak{}binomial summary tables. No claim of an executed analysis artifact is made at this stage.\par
2. Use only the 90 normalized expression columns, not the undocumented comparison blocks. Remove the 2,210 globally all-\allowbreak{}zero genes; retain other genes with estimable positive residual variance for each contrast, recording exact denominators and exclusions. Do not invent low-\allowbreak{}expression or outlier thresholds.\par
3. For each requested pair (Final minus Baseline Blood; Dentate gyrus minus Baseline Blood; Dentate gyrus minus Final Blood), fit a gene-\allowbreak{}wise linear model to log2(normalized count + 1), using tissue plus categorical Response. Use heteroskedasticity-\allowbreak{}robust HC3 coefficient tests with residual-\allowbreak{}degrees-\allowbreak{}of-\allowbreak{}freedom t calibration. This is a feasible approximate analysis of continuous normalized data, not a reconstruction of the workbook's DESeq analysis. Report coefficients and both nominal p-\allowbreak{}values and Benjamini–Hochberg q-\allowbreak{}values within each contrast. A tissue-\allowbreak{}by-\allowbreak{}response model/\allowbreak{}equal-\allowbreak{}response-\allowbreak{}average contrast can serve as a sensitivity check if response heterogeneity matters. Pairing will not be introduced absent explicit approval of a clearly labelled assumption.\par
4. Precisely specified proposed binomial interpretation: for each contrast, each tested gene is one trial (n); a success is a nominal, two-\allowbreak{}sided gene-\allowbreak{}level p-\allowbreak{}value <0.05 (k); under the global null the success probability is p0=\allowbreak{}0.05. Compute scipy.stats.binomtest(k,n,p=\allowbreak{}0.05,alternative=\allowbreak{}'greater') to test enrichment of nominal differential signals. Report k, n, k/\allowbreak{}n, and exact-\allowbreak{}test p-\allowbreak{}values. Apply Holm correction across the three binomial p-\allowbreak{}values, judging evidence of enrichment at adjusted p<0.05; retain unadjusted values for transparency. Do not count BH-\allowbreak{}significant genes as Bernoulli(0.05) successes, because that would not have this null probability. BH DEG counts are supplementary, separate summaries.\par
\smallskip
INTERPRETATION AND LIMITATIONS\par
The task specifies a binomial exact test and a 5\% threshold, but not the trial definition, null probability, direction, or multiplicity convention. The above is an explicit proposal, not a sourced original protocol. 'Negligible effects' could mean no excess nominal discoveries, few FDR discoveries, small expression changes, or formal biological equivalence. Under the proposal, lack of enrichment means only insufficient evidence of excess differential signal; it does not prove negligible effect sizes. I would compare discovery fractions and effect-\allowbreak{}size distributions, and avoid an equivalence claim without a prespecified biological margin.\par
\smallskip
The binomial calculation is exact only under independent Bernoulli trials with the specified probability. Correlated genes and approximate gene-\allowbreak{}level p-\allowbreak{}values violate that idealization; its p-\allowbreak{}value must be labelled a heuristic global summary. Samples may also be repeated measurements, making unpaired-\allowbreak{}model inference uncertain. Normalized fractional values cannot legitimately be rounded and passed off as original DESeq/\allowbreak{}negative-\allowbreak{}binomial counts; raw counts, size factors, normalization details, and dispersion/\allowbreak{}outlier settings are absent. Read totals do not reconstruct raw gene counts. Very large counts, substantial sparsity, and tissue composition differences warrant diagnostics. No tissue comparison can yet be selected as the answer from inspection alone.\par
\interactionfield{Stage}
plan\par
\interactionfield{Evidence}
Observed 90/\allowbreak{}90 unique Code-\allowbreak{}header matches; tissue × Response table has nine cells of 10 samples each. NormCount has 25,402 unique genes and 2,210 all-\allowbreak{}zero rows, leaving at most 23,192 genes before contrast-\allowbreak{}specific estimability exclusions. Workbook includes nine numbered DESeq result blocks without contrast definitions, and no explicit subject or batch column. Inspection executed via openpyxl/\allowbreak{}NumPy only; no inferential analyses were run.\par
\interactionfield{Assumptions}
[\par
  "Proposed log2(count+1) analysis treats available values as normalized continuous expression, not raw counts.",\par
  "Primary unpaired modeling would assume independent samples because subject identity is unverified.",\par
  "Response adjustment, nominal-\allowbreak{}gene-\allowbreak{}p-\allowbreak{}value binomial successes, one-\allowbreak{}sided enrichment, and Holm correction are proposed choices requiring approval."\par
]\par
\interactionfield{Uncertainties}
[\par
  "Subject pairing and biological meanings of filename tokens are not documented.",\par
  "Normalization algorithm, raw counts, original outlier exclusions and numbered DESeq contrast definitions are unavailable.",\par
  "'Negligible effects' and the intended binomial trial/\allowbreak{}null construction are not defined by supplied documentation."\par
]\par
\end{interactionevent}
\begin{interactionevent}{Event 11: Researcher (simulator)}
\interactionfield{Message}
Approved: execute the proposed response-\allowbreak{}adjusted log2(normalized count + 1) HC3 analysis for all three contrasts, with nominal p<0.05 successes in the one-\allowbreak{}sided exact binomial enrichment test against p0=\allowbreak{}0.05, and report both raw and Holm-\allowbreak{}adjusted binomial p-\allowbreak{}values. These are explicit working choices, not recovered source methods. Keep BH discovery counts separate. Do not infer pairing or assign the undocumented DESeq blocks to contrasts. Include the proposed tissue-\allowbreak{}by-\allowbreak{}response sensitivity analysis to assess whether averaging responses masks heterogeneous effects. Record tested-\allowbreak{}gene denominators, variance/\allowbreak{}estimability exclusions, discovery counts and fractions, and effect-\allowbreak{}size summaries alongside basic model diagnostics. Interpret lack of enrichment as lack of evidence, not equivalence; if all contrasts show enrichment, assess relative effect magnitude without calling a nonsignificant global test that did not occur. Provide the concise best-\allowbreak{}supported answer to which comparison shows negligible effects, noting consequential uncertainty if needed, and retain the executed script at analysis/\allowbreak{}tissue\_\allowbreak{}differential\_\allowbreak{}expression.py (or report its actual path). Final submission must include stage=\allowbreak{}'final', answer, payload.executed\_\allowbreak{}analysis=\allowbreak{}true only after execution, and artifacts.analysis pointing to the executed script or notebook.\par
\interactionfield{Interaction}
next\_\allowbreak{}task\par
\interactionfield{Phase}
approach\_\allowbreak{}review\par
\interactionfield{Action}
continue\par
\interactionfield{Approve Approach}
true\par
\interactionfield{Evidence}
Inspection reports a complete 90-\allowbreak{}sample metadata join, balanced tissue-\allowbreak{}by-\allowbreak{}response groups, normalized fractional expression and 23,192 non-\allowbreak{}all-\allowbreak{}zero genes. Raw counts, pairing, original contrast definitions and binomial specifications are unavailable. The proposed method is executable and transparently distinguishes assumptions from documented input properties.\par
\interactionfield{Objectives Open}
[\par
  "Execute the three differential-\allowbreak{}expression contrasts and exact binomial tests with reported diagnostics.",\par
  "Evaluate response heterogeneity and identify the comparison with negligible relative effects without claiming formal equivalence.",\par
  "Return the concise answer and executed analysis artifact under the submission contract."\par
]\par
\end{interactionevent}
\begin{interactionevent}{Event 18: Actor}
\interactionfield{Answer}
Final vs Baseline Blood shows the smallest expression effects relative to either dentate-\allowbreak{}gyrus comparison. However, none of the three comparisons shows negligible effects in the sense of lacking binomial enrichment: all are strongly enriched for differential-\allowbreak{}expression signals. Thus, Final vs Baseline Blood is the best-\allowbreak{}supported relative answer, not evidence of biological equivalence.\par
\interactionfield{Stage}
result\par
\interactionfield{Evidence}
Execution completed successfully (exit 0), including a rerun using stable log-\allowbreak{}summed binomial tails to resolve floating-\allowbreak{}point underflow. Six gene-\allowbreak{}level CSV files, summary.json, and execution.log were produced by the retained script.\par
\smallskip
PRIMARY RESPONSE-\allowbreak{}ADJUSTED HC3 RESULTS\par
Contrast | Tested genes | Additional exclusions after 2,210 global all-\allowbreak{}zero genes | Nominal p<0.05 successes | BH q<0.05 discoveries | Median absolute log2(count+1) effect | Fraction absolute effect >1\par
Final–Baseline Blood | 22,060 | 1,132 | 11,083 (50.24\%) | 9,989 (45.28\%) | 0.217 | 3.91\%\par
Dentate–Baseline Blood | 23,017 | 175 | 19,526 (84.83\%) | 19,403 (84.30\%) | 1.179 | 54.81\%\par
Dentate–Final Blood | 22,890 | 302 | 19,305 (84.34\%) | 19,165 (83.73\%) | 1.208 | 55.38\%\par
\smallskip
Additional exclusions were genes with residual sum of squares or HC3 contrast variance <=\allowbreak{}1e-\allowbreak{}20, or nonfinite variance. Every pair contains 60 samples, balanced across tissue and response groups. BH correction was performed separately within each contrast and never used to define binomial successes.\par
\smallskip
BINOMIAL EXACT TEST\par
Trials =\allowbreak{} tested genes; successes =\allowbreak{} two-\allowbreak{}sided HC3 nominal p<0.05; null probability =\allowbreak{} 0.05; alternative =\allowbreak{} greater. Holm adjustment covers the three primary contrast tests. Raw and Holm-\allowbreak{}adjusted log10(p), respectively:\par
• Final–Baseline: −8025.479, −8025.479.\par
• Dentate–Baseline: −21229.487, −21229.010.\par
• Dentate–Final: −20883.787, −20883.486.\par
Ordinary floating-\allowbreak{}point exact-\allowbreak{}test p-\allowbreak{}values underflow to zero; these log-\allowbreak{}scale tail sums represent extremely small positive probabilities, not literal zero. All reject enrichment nulls at 5\%, before and after Holm adjustment. No nonsignificant global test occurred.\par
\smallskip
EFFECT-\allowbreak{}SIZE DISTRIBUTIONS\par
Absolute-\allowbreak{}effect quartiles (25th, 50th, 75th percentiles) are 0.086/\allowbreak{}0.217/\allowbreak{}0.431 for Final–Baseline, 0.395/\allowbreak{}1.179/\allowbreak{}2.865 for Dentate–Baseline, and 0.401/\allowbreak{}1.208/\allowbreak{}2.973 for Dentate–Final. Their 95th percentiles are 0.935, 7.361, and 7.509. These are differences of mean log2(count+1), not exact raw-\allowbreak{}count log fold changes.\par
\smallskip
TISSUE-\allowbreak{}BY-\allowbreak{}RESPONSE SENSITIVITY\par
The interaction model estimates an equally weighted average of the three response-\allowbreak{}specific tissue effects. Balanced sampling makes those average estimates identical to the additive estimates, but standard errors differ. Tested denominators/\allowbreak{}exclusions are unchanged. Nominal discoveries are 10,966 (49.71\%), 19,483 (84.65\%), and 19,253 (84.11\%); BH discoveries are 9,830 (44.56\%), 19,380 (84.20\%), and 19,129 (83.57\%), respectively. Raw/\allowbreak{}Holm log10 binomial p-\allowbreak{}values are −7875.915/\allowbreak{}−7875.915, −21142.488/\allowbreak{}−21142.011, and −20779.466/\allowbreak{}−20779.165 (Holm applied separately to sensitivity tests).\par
For Final–Baseline, response-\allowbreak{}specific median absolute effects are 0.202 Control, 0.302 good\_\allowbreak{}responder, and 0.298 bad\_\allowbreak{}responder. Response-\allowbreak{}specific tested/\allowbreak{}nominal/\allowbreak{}BH counts are 20,956/\allowbreak{}6,014/\allowbreak{}3,831; 20,783/\allowbreak{}7,962/\allowbreak{}6,208; and 20,879/\allowbreak{}9,155/\allowbreak{}7,751. Thus averaging does not conceal an otherwise absent blood effect; responder effects are somewhat larger. Dentate comparison response-\allowbreak{}specific medians remain 1.228–1.300. Across the 23,192 globally nonzero genes, 30.43\% have mixed response-\allowbreak{}specific effect signs in the blood contrast (often small effects), versus 7.79\% and 8.57\% in dentate contrasts. Only 0.418\% have blood average absolute effect <0.25 while some group has absolute effect >1, versus 0\% and 0.0345\% for dentate contrasts. These descriptive thresholds assess cancellation, not a prespecified equivalence margin or a formal interaction test.\par
\smallskip
BASIC MODEL DIAGNOSTICS\par
All primary designs have rank 4/\allowbreak{}4, residual df=\allowbreak{}56, condition number 4.22, and uniform leverage 0.0667. Interaction designs have rank 6/\allowbreak{}6, df=\allowbreak{}54, condition number 9.77, and leverage 0.10. Median residual variances on the transformed scale are 0.124, 0.120, and 0.141 for primary models; no rank deficiency was found. Full quantiles and group-\allowbreak{}specific sensitivity results are retained in summary.json. No inferred pairing, undocumented DESeq contrast assignment, or invented outlier exclusions were used.\par
\interactionfield{Assumptions}
[\par
  "Normalized expression was modeled as log2(count+1) with categorical Response adjustment and HC3 standard errors, using residual-\allowbreak{}df t calibration; these are approved working methods, not recovered original methods.",\par
  "Samples were treated as independent because subject pairing is undocumented.",\par
  "Binomial calculations assume independent, correctly calibrated gene-\allowbreak{}level null trials; actual gene dependence means the exact binomial tail is only a heuristic global summary."\par
]\par
\interactionfield{Uncertainties}
[\par
  "Raw gene counts, normalization/\allowbreak{}size-\allowbreak{}factor provenance, and original DESeq outlier rules are absent; the analysis is approximate and does not reconstruct DESeq.",\par
  "Potential repeated measurements could change inference, but subject identities were not inferred from codes.",\par
  "No biological equivalence margin defines 'negligible'; the blood comparison has smaller effects but substantial differential expression, so an unqualified no-\allowbreak{}effect claim would contradict the executed results."\par
]\par
\end{interactionevent}
\begin{interactionevent}{Event 19: Researcher (simulator)}
\interactionfield{Message}
The reported analysis adequately addresses the agreed approach. Submit the final candidate identifying Final vs Baseline Blood as showing the smallest relative effects, while retaining the qualification that all three contrasts show significant binomial enrichment and therefore none supports an unqualified absence-\allowbreak{}of-\allowbreak{}effect conclusion. Include stage=\allowbreak{}'final', the concise answer, payload.executed\_\allowbreak{}analysis=\allowbreak{}true, and artifacts.analysis=\allowbreak{}'analysis/\allowbreak{}tissue\_\allowbreak{}differential\_\allowbreak{}expression.py'.\par
\interactionfield{Interaction}
conclude\par
\interactionfield{Phase}
result\_\allowbreak{}review\par
\interactionfield{Action}
stop\par
\interactionfield{Approve Approach}
false\par
\interactionfield{Evidence}
Reported execution includes all three contrasts, response-\allowbreak{}adjusted HC3 tests, exact binomial tails with Holm correction, separate BH counts, interaction sensitivity analysis, and estimability diagnostics. The blood contrast has median absolute transformed effect 0.217 versus 1.179 and 1.208 for dentate contrasts, but 50.24\% nominal discoveries and strong binomial enrichment. The candidate preserves this distinction and supplies the required execution artifact; no identifiable mismatch warrants revision.\par
\interactionfield{Objectives Open}
[]\par
\end{interactionevent}
\Needspace{10\baselineskip}
\subsection{Incremental Clarification: SciConvBench with Persona 127}
\label{app:case-clarification}
\begin{figure}[!htbp]
\centering
\begingroup
\definecolor{casehighlight}{RGB}{255,241,173}
\sethlcolor{casehighlight}
\begin{tcolorbox}[colback=gray!3,colframe=gray!55,boxrule=0.5pt,
  arc=1pt,left=6pt,right=6pt,top=6pt,bottom=6pt,
  fontupper=\footnotesize]
\textbf{Incremental clarification (Case 2)}\par
\textit{SciConvBench; persona 127.}\par
\textbf{Actor, event 2:} What applied torque should be used to size the shaft?\par
\textbf{Researcher, event 3:} Use an applied torque of
$T=1{,}000{,}000\,\mathrm{N\!\cdot\!mm}$ to size the shaft.\par
\textbf{Actor, event 6:} What is the maximum allowable angle of twist over the full
shaft length?\par
\textbf{Researcher, event 7:} The maximum allowable angle of twist is
$\phi_{\mathrm{allow}}=0.75^\circ$ over the full 1500\,mm shaft length.
\hl{This is approximately 0.01309 radians for use in the torsion equation.}\par
\textbf{Researcher, event 11:} [\ldots] \hl{Size the shaft solely from the
angle-of-twist limit using $\phi=TL/(JG)$ and $J=\pi d^4/32$.}\par
\textit{Five clarification exchanges in total; in the matched persona-off
conversation, one reply lists every parameter.}\par
\textit{Full interactions: Appendices~\ref{app:case-clarification}
and~\ref{app:case-clarification-off}.}
\end{tcolorbox}
\endgroup
\caption{\textbf{Incremental clarification with explanatory replies.}
Highlights added; [\ldots] marks omissions. Event identifiers are preserved.}
\label{fig:discussion-clarification}
\end{figure}

Both SciConvBench conversations received full credit despite different
clarification paths (Figure~\ref{fig:discussion-clarification}).
Task: \texttt{solMech/\allowbreak{}disambiguation/\allowbreak{}case\_\allowbreak{}080}. Trial 1; persona 127.
\begin{interactionevent}{Event 1: Initial task}
\interactionfield{Message}
Assume small-\allowbreak{}deformation, linear-\allowbreak{}elastic solid mechanics unless otherwise required.\par
\smallskip
A solid circular shaft is subjected to a torque and shear loading.\par
\smallskip
Determine the relevant response quantities for the case, focusing on: design a solid shaft diameter from an angle-\allowbreak{}of-\allowbreak{}twist limit.\par
\smallskip
State any additional assumptions you need and present the governing relations clearly.\par
\end{interactionevent}
\begin{interactionevent}{Event 2: Actor clarification question}
\interactionfield{Question}
What applied torque should be used to size the shaft?\par
\end{interactionevent}
\begin{interactionevent}{Event 3: Researcher (simulator) reply}
\interactionfield{Answer}
Use an applied torque of **T =\allowbreak{} 1,000,000 N·mm** to size the shaft.\par
\end{interactionevent}
\begin{interactionevent}{Event 4: Actor clarification question}
\interactionfield{Question}
What is the shaft length over which the angle-\allowbreak{}of-\allowbreak{}twist limit applies?\par
\end{interactionevent}
\begin{interactionevent}{Event 5: Researcher (simulator) reply}
\interactionfield{Answer}
The angle-\allowbreak{}of-\allowbreak{}twist limit applies over the full shaft length, **L =\allowbreak{} 1500 mm**.\par
\end{interactionevent}
\begin{interactionevent}{Event 6: Actor clarification question}
\interactionfield{Question}
What is the maximum allowable angle of twist over the full shaft length?\par
\end{interactionevent}
\begin{interactionevent}{Event 7: Researcher (simulator) reply}
\interactionfield{Answer}
The maximum allowable angle of twist is **φ\_\allowbreak{}allow =\allowbreak{} 0.75°** over the full **1500 mm** shaft length. This is approximately **0.01309 radians** for use in the torsion equation.\par
\end{interactionevent}
\begin{interactionevent}{Event 8: Actor clarification question}
\interactionfield{Question}
What is the shear modulus G of the shaft material?\par
\end{interactionevent}
\begin{interactionevent}{Event 9: Researcher (simulator) reply}
\interactionfield{Answer}
The shaft material is **6061-\allowbreak{}T6 aluminum**, with shear modulus **G =\allowbreak{} 25,939.8 MPa** (equivalently, **25,939.8 N/\allowbreak{}mm²**).\par
\end{interactionevent}
\begin{interactionevent}{Event 10: Actor clarification question}
\interactionfield{Question}
Does “shear loading” refer only to the shear stress induced by the torque, or is there an additional transverse shear force?\par
\end{interactionevent}
\begin{interactionevent}{Event 11: Researcher (simulator) reply}
\interactionfield{Answer}
“Shear loading” refers only to the shear stress induced by the applied torque. No additional transverse shear force is specified. Size the shaft solely from the angle-\allowbreak{}of-\allowbreak{}twist limit using \textbackslash{}(\textbackslash{}phi =\allowbreak{} TL/\allowbreak{}(JG)\textbackslash{}) and \textbackslash{}(J =\allowbreak{} \textbackslash{}pi d\textasciicircum{}4/\allowbreak{}32\textbackslash{}).\par
\end{interactionevent}
\begin{interactionevent}{Event 12: Actor final specification}
\interactionfield{Answer}
Determine the minimum diameter of a solid circular shaft made from 6061-\allowbreak{}T6 aluminum, sizing solely from the allowable angle of twist.\par
Use an applied torque T =\allowbreak{} 1,000,000 N·mm, shaft length L =\allowbreak{} 1500 mm, and shear modulus G =\allowbreak{} 25,939.8 N/\allowbreak{}mm².\par
Use a maximum allowable twist over the full length of φ\_\allowbreak{}allow =\allowbreak{} 0.75° =\allowbreak{} π/\allowbreak{}240 radians, approximately 0.01309 radians.\par
Assume small-\allowbreak{}deformation, linear-\allowbreak{}elastic behavior, a homogeneous isotropic material, a uniform circular cross-\allowbreak{}section, and constant internal torque along the shaft.\par
Assume the applied torque is balanced by a reaction torque and neglect localized end effects.\par
Treat shear loading as torsional shear stress only, with no additional transverse shear force.\par
Present the governing relations φ =\allowbreak{} TL/\allowbreak{}(JG) and J =\allowbreak{} πd⁴/\allowbreak{}32.\par
Calculate the required polar second moment of area from J\_\allowbreak{}required =\allowbreak{} TL/\allowbreak{}(Gφ\_\allowbreak{}allow).\par
Calculate the minimum shaft diameter from d\_\allowbreak{}min =\allowbreak{} [32TL/\allowbreak{}(πGφ\_\allowbreak{}allow)]\textasciicircum{}(1/\allowbreak{}4), using radians for the twist limit.\par
Report the calculated minimum diameter in millimeters without rounding downward.\par
Verify that the twist at the reported diameter does not exceed the allowable limit.\par
Report the maximum torsional shear stress at the shaft surface using τ\_\allowbreak{}max =\allowbreak{} T(d/\allowbreak{}2)/\allowbreak{}J =\allowbreak{} 16T/\allowbreak{}(πd³).\par
Do not impose an additional strength criterion or claim strength adequacy, since sizing is solely from the twist limit and no allowable stress is specified.\par
\end{interactionevent}
\Needspace{10\baselineskip}
\subsection{Matched SciConvBench Conversation without a Persona}
\label{app:case-clarification-off}
Task: \texttt{solMech/\allowbreak{}disambiguation/\allowbreak{}case\_\allowbreak{}080}. Trial 1; persona persona\_\allowbreak{}off.
\begin{interactionevent}{Event 1: Initial task}
\interactionfield{Message}
Assume small-\allowbreak{}deformation, linear-\allowbreak{}elastic solid mechanics unless otherwise required.\par
\smallskip
A solid circular shaft is subjected to a torque and shear loading.\par
\smallskip
Determine the relevant response quantities for the case, focusing on: design a solid shaft diameter from an angle-\allowbreak{}of-\allowbreak{}twist limit.\par
\smallskip
State any additional assumptions you need and present the governing relations clearly.\par
\end{interactionevent}
\begin{interactionevent}{Event 2: Actor clarification question}
\interactionfield{Question}
Should the shaft diameter be derived symbolically, or calculated numerically from specified loading, geometry, material properties, and twist limit?\par
\end{interactionevent}
\begin{interactionevent}{Event 3: Researcher (simulator) reply}
\interactionfield{Answer}
Please calculate the diameter numerically using N–mm–MPa units:\par
\smallskip
-\allowbreak{} Length: \textbackslash{}(L =\allowbreak{} 1500\textbackslash{} \textbackslash{}text\{mm\}\textbackslash{})\par
-\allowbreak{} Material: 6061-\allowbreak{}T6 aluminum\par
-\allowbreak{} Shear modulus: \textbackslash{}(G =\allowbreak{} 25939.8\textbackslash{} \textbackslash{}text\{MPa\}\textbackslash{})\par
-\allowbreak{} Torque: \textbackslash{}(T =\allowbreak{} 1\{,\}000\{,\}000\textbackslash{} \textbackslash{}text\{N·mm\}\textbackslash{})\par
-\allowbreak{} Allowable twist: \textbackslash{}(\textbackslash{}phi\_\allowbreak{}\{\textbackslash{}text\{allow\}\} =\allowbreak{} 0.75\textasciicircum{}\textbackslash{}circ\textbackslash{}) over the full length\par
\smallskip
Determine the minimum solid-\allowbreak{}shaft diameter using \textbackslash{}(\textbackslash{}phi =\allowbreak{} TL/\allowbreak{}(JG)\textbackslash{}) and \textbackslash{}(J =\allowbreak{} \textbackslash{}pi d\textasciicircum{}4/\allowbreak{}32\textbackslash{}), converting the twist limit to radians. Please also show the governing expression for the diameter.\par
\end{interactionevent}
\begin{interactionevent}{Event 4: Actor final specification}
\interactionfield{Answer}
Determine the minimum solid circular shaft diameter numerically using N–mm–MPa units.\par
Assume small-\allowbreak{}deformation, linear-\allowbreak{}elastic behavior for a homogeneous, isotropic 6061-\allowbreak{}T6 aluminum shaft with constant circular cross section.\par
Use a shaft length L =\allowbreak{} 1500 mm, shear modulus G =\allowbreak{} 25939.8 MPa, and constant internal torque T =\allowbreak{} 1,000,000 N·mm over the full length.\par
Limit the relative angle of twist over the full shaft length to 0.75 degrees.\par
Convert the allowable twist to radians using φ\_\allowbreak{}allow =\allowbreak{} 0.75π/\allowbreak{}180.\par
Use the governing torsion relations φ =\allowbreak{} TL/\allowbreak{}(JG) and J =\allowbreak{} πd⁴/\allowbreak{}32.\par
Show that the twist constraint gives d ≥ [32TL/\allowbreak{}(πGφ\_\allowbreak{}allow)]\textasciicircum{}(1/\allowbreak{}4).\par
Substitute the numerical values and report the minimum diameter in millimeters without rounding downward.\par
Verify that the reported diameter satisfies the allowable twist.\par
Treat this as sizing solely for torsional stiffness; no separate transverse shear load is specified.\par
State that strength, bending, transverse shear, stress concentrations, and connection effects are outside this sizing calculation.\par
\end{interactionevent}

\end{document}